\documentclass[fleqn]{SCYE}

\usepackage{xcolor}
\usepackage{color}

\begin{document}

%%%%%%%%%%%%%%%%%%%%%%%%%%%%%%%%%%%%%%%%%%%%%%%%%%%%%%%
%%% Authors do not modify the information below
%%% ????????????????
%%% ??????????, ????????????{}, ???????????????????
%Letter to the Editor??Article%??????
\ArticleType{Invited Review}%??Article
%\SpecialTopic{SPECIAL TOPIC: natural language processing}%???????
%\Year{2020}
%\Month{}
%\Vol{}
%\No{}
%\DOI{}
%\BeginPage{} % 起页码
%\EndPage{}
%\ReceiveDate{}
%\AcceptDate{}
%\OnlineDate{}
%%%%%%%%%%%%%%%%%%%%%%%%%%%%%%%%%%%%%%%%%%%%%%%%%%%%%%%

%%% title: ????
%%%   \title{title}{title for citation}
\title{Neural Machine Translation: Challenges, Progress and Future}
{Neural Machine Translation: Challenges, Progress and Future}

%%% Corresponding author: ???????
%%%   \author[number]{Full name}{{email@xxx.com}}
%%% General author: ???????
%%%   \author[number]{Full name}{}
\author[1,2]{Jiajun Zhang}{{jjzhang@nlpr.ia.ac.cn}}%
\author[1,2,3]{Chengqing Zong}{{cqzong@nlpr.ia.ac.cn}}

%%% Author information for page head. ?ü?е????????
%%% ??????????????, ??????????author???
\AuthorMark{Zhang JJ and Zong CQ}%\authorcr????????

%%% Authors for citation. ????????е????????
%%% ??????????????, ??????????author???
\AuthorCitation{Zhang JJ and Zong CQ.}

%%% Address.
%%%   \address[number]{Address, City {\rm Postcode}, Country}
\address[1]{National Laboratory of Pattern Recognition, CASIA, Beijing, China}
\address[2]{School of Artificial Intelligence, University of Chinese Academy of Sciences, Beijing, China}
\address[3]{CAS Center for Excellence in Brain Science and Intelligence Technology, Shanghai, China }

%\contributions{These authors contributed equally to the work.}%

%%% Abstract.
\abstract{Machine translation (MT) is a technique that leverages computers to translate human languages automatically. Nowadays, neural machine translation (NMT) which models direct mapping between source and target languages with deep neural networks has achieved a big breakthrough in translation performance and become the {\em de facto} paradigm of MT. This article makes a review of NMT framework, discusses the challenges in NMT, introduces some exciting recent progresses and finally looks forward to some potential future research trends. In addition, we maintain the state-of-the-art methods for various NMT tasks at the website https://github.com/ZNLP/SOTA-MT.}

%%% Keywords.
\keywords{neural machine translation, Transformer, multimodal translation, low-resource translation, document translation}

\maketitle

%\tableofcontents%?????

%%%%%%%%%%%%%%%%%%%%%%%%%%%%%%%%%%%%%%%%%%%%%%%%%%%%%%%
%%% The main text. ???????
%\cite{3,4,5,6}  链用 [1－3]
% 公式引用        自动加括号
%\eqref{eq1}   (1)
%\cref{eq1}     eq.(1)
%\Cref{eq1}    Eq.(1)
%\cref{fig1}   Figure 1
%\cref{tab1}   Table  1
%\href{链接网址}{显示网址}%%\href{https://mc03. manuscriptcentral.com/scpma}{https://mc03.manuscriptcentral.com/scpma}

%\twocolumn\onecolumn
%%%%%%%%%%%%%%%%%%%%%%%%%%%%%%%%%%%%%%%%%%%%%%%%%%%%%%%
\begin{multicols}{2}

%\Authorfootnote

\section{Introduction}\label{section1}
The concept of machine translation (MT) was formally proposed in 1949 by Warren Weaver \cite{weaver1955translation} who believed it is possible to use modern computers to automatically translate human languages. From then on, machine translation has become one of the most challenging task in the area of natural language processing and artificial intelligence. Many researchers of several generations dedicated themselves to realize the dream of machine translation.

From the viewpoint of methodology, approaches to MT mainly fall into two categories: rule-based method and data-driven approach. Rule-based methods were dominant and preferable before 2000s. In this kind of methods, bilingual linguistic experts are responsible to design specific rules for source language analysis, source-to-target language transformation and target language generation. Since it is very subjective and labor intensive, rule-based systems are difficult to be scalable and they are fragile when rules cannot cover the unseen language phenomena.

In contrast, the data-driven approach aims at teaching computers to learn how to translate from lots of human-translated parallel sentence pairs (parallel corpus). The study of data-driven approach has experienced three periods. In the middle of 1980s, \cite{nagao1984framework} proposed example-based MT which translates a sentence by retrieving the similar examples in the human-translated sentence pairs. 

From early 1990s, statistical machine translation (SMT) has been proposed and word or phrase level translation rules can be automatically learned from parallel corpora using probabilistic models \cite{brown1993mathematics,koehn2003statistical,chiang2005hierarchical}. Thanks to the availability of more and more parallel corpora, sophisticated probabilistic models such as noisy channel model and log-linear model achieve better and better translation performance. Many companies (e.g. Google, Microsoft and Baidu) have developed online SMT systems which much benefit the users. However, due to complicated integration of multiple manually designed components such as translation model, language model and reordering model, SMT cannot make full use of large-scale parallel corpora and translation quality is far from satisfactory. 

No breakthrough has been achieved more than 10 years until the introduction of deep learning into MT. Since 2014, neural machine translation (NMT) based on deep neural networks has quickly developed \cite{sutskever2014sequence,bahdanau2015neural,gehring2017convolutional,vaswani2017attention}. In 2016, through extensive experiments on various language pairs, \cite{junczys2016neural,wu2016google} demonstrated that NMT has made a big breakthrough and obtained remarkable improvements compared to SMT, and even approached human-level translation quality \cite{hassan2018achieving}. This article attempts to give a review of NMT framework, discusses some challenging research tasks in NMT, introduces some exciting progresses and forecasts several future research topics.

The remainder of this article is organized as follows: Sec.~\ref{section2} first introduces the background and state-of-the-art paradigm of NMT. In Sec.~\ref{section3} we discuss the key challenging research tasks in NMT. From Sec.~\ref{section4} to Sec.~\ref{section7}, the recent progresses are presented concerning each challenge. Sec.~\ref{section8} discusses the current state of NMT compared to expert translators and finally looks forward to some potential research trends in the future.

\section{Neural machine translation}\label{section2}
\subsection{Encoder-Decoder Framework}\label{subsection21}
Neural machine translation is an end-to-end model following an encoder-decoder framework that usually includes two neural networks for encoder and decoder respectively \cite{sutskever2014sequence,bahdanau2015neural,gehring2017convolutional,vaswani2017attention}. As shown in Fig.~\ref{fig1}, the {\em encoder} network first maps each input token of the source-language sentence into a low-dimensional real-valued vector (aka word embedding) and then encodes the sequence of vectors into distributed semantic representations, from which the {\em decoder} network generates the target-language sentence token by token {\footnote{Currently, subword is the most popular translation token for NMT \cite{sennrich2016neural}.}} from left to right.

From the probabilistic perspective, NMT models the conditional probability of the target-language sentence ${\bm y}=y_0, \cdots, y_i, \cdots, y_I$ given the source-language sentence ${\bm x}=x_0,\cdots,x_j,\cdots, x_J$ as a product of token-level translation probabilities.

\begin{equation}
P({\bm y}|{\bm x},\theta)=\prod_{i=0}^{I}P(y_i|{\bm x},{\bm y}_{<i},\theta)
\end{equation}
where ${\bm y}_{<i}=y_0, \cdots, y_{i-1}$ is the partial translation which has been generated so far. $x_0$, $y_0$ and $x_J$, $y_I$ are often special symbols $<$s$>$ and $<$/s$>$ indicating the start and end of a sentence respectively.

The token-level translation probability can defined as follows:

\begin{figure}[H]
	\centering
	\includegraphics[scale=.6]{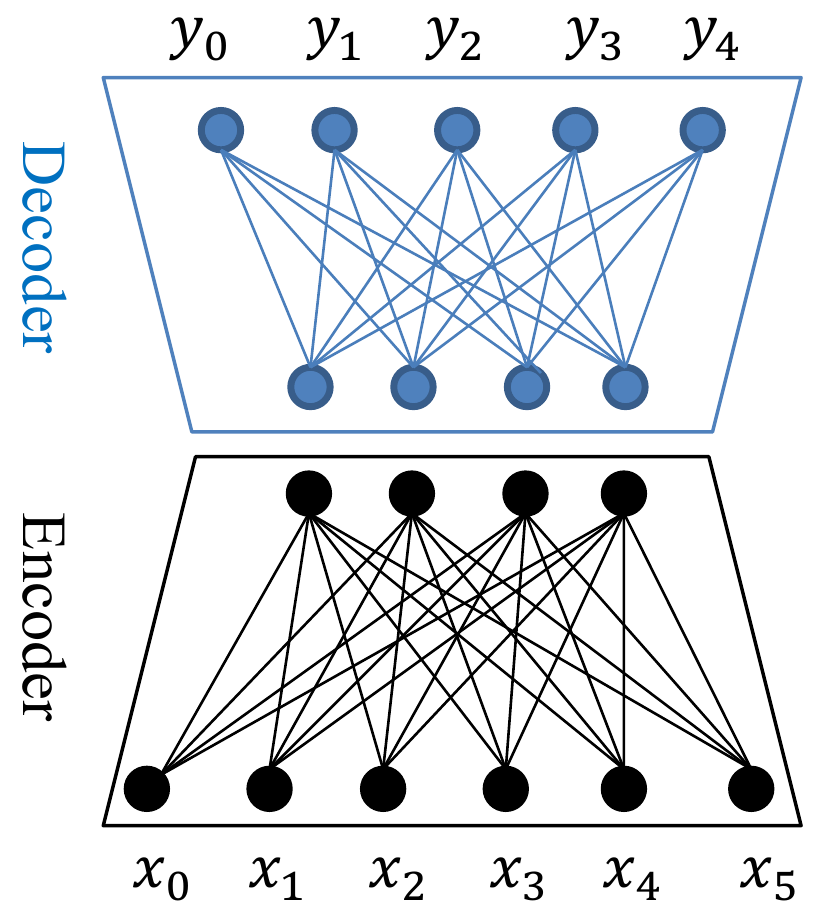}
	\caption{encoder-decoder framework for neural machine translation. The encoder encodes the input sequence $x_0x_1x_2x_3x_4x_5$ into distributed semantic representations based on which the decoder produces an output sequence $y_0y_1y_2y_3y_4$.}
	\label{fig1}
\end{figure}

\begin{equation}
P(y_i|{\bm x},{\bm y}_{<i},\theta)=\frac{{\text{exp}}{\bigg(}g({{\bm x}, {\bm y}_{<i}, y_i, \theta}){\bigg)}}{\sum_{y' \in V}{\text{exp}}{\bigg(}g({{\bm x}, {\bm y}_{<i}, y', \theta}){\bigg)}}
\end{equation}
in which $V$ denotes the vocabulary of the target language and $g(\cdot)$ is a non-linear function that calculates a real-valued score for the prediction $y_i$ conditioned on the input ${\bm x}$, the partial translation ${\bm y}_{<i}$ and the model parameters $\theta$. The non-linear function $g(\cdot)$ is realized through the encoder and decoder networks. The input sentence ${\bm x}$ is abstracted into hidden semantic representations ${\bm h}$ through multiple encoder layers. ${\bm y}_{<i}$ is summarized into the target-side history context representation ${\bm z}$ with decoder network which further combines ${\bm h}$ and ${\bm z}$ using an attention mechanism to predict the score of $y_i$.

The network parameters $\theta$ can be optimized to maximize the log-likelihood over the bilingual training data $D=\{({\bm{x}}^{(m)},{\bm{y}}^{(m)})\}_{m=1}^M$:

\begin{equation}
\theta = argmax_{\theta^*} \sum_{m=1}^{M} {\text{log}}P({\bm y}^{(m)}|{\bm x}^{(m)},\theta^*)
\end{equation}

These years have witnessed the fast development of the encoder-decoder networks from recurrent neural network \cite{sutskever2014sequence,bahdanau2015neural}, to convolutional neural network \cite{gehring2017convolutional} and then to self-attention based neural network Transformer \cite{vaswani2017attention}. At present, Transformer is the state-of-the-art in terms of both quality and efficiency.

\subsection{Transformer}\label{subsection22}

\begin{figure}[H]
	\centering
	\includegraphics[scale=.5]{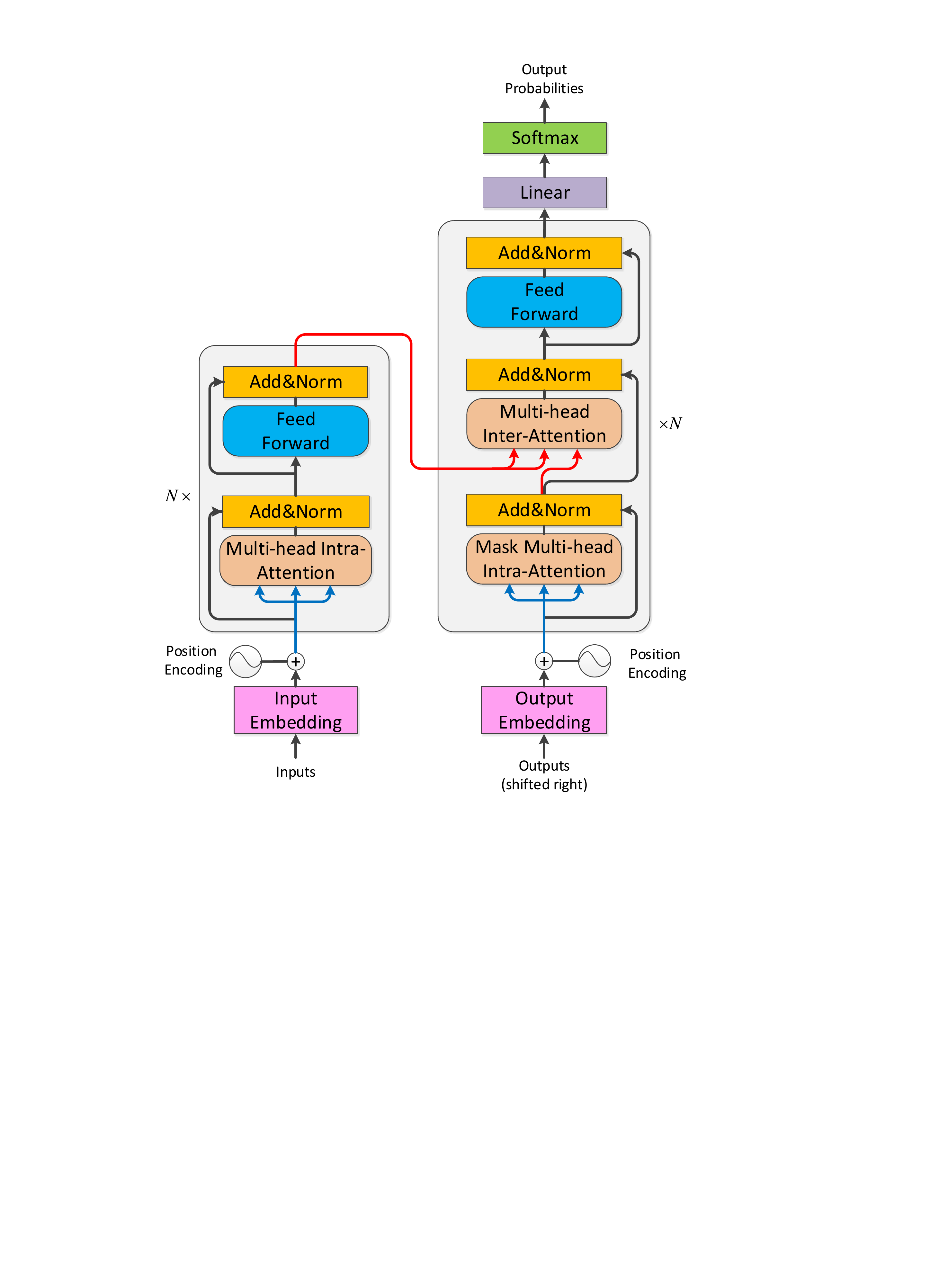}
	\caption{the Transformer architecture in which attention mechanism is the core in both of the encoder and decoder networks. {\em{shift right}} means that the prediction of the previous time-step will shift right as the input context to predict next output token.}
	\label{fig2}
\end{figure}

In Transformer{\footnote{Model and codes can be found at https://github.com/tensorflow/tensor2tensor}}, the encoder includes $N$ identical layers and each layer is composed of two sub-layers: the self-attention sub-layer followed by the feed-forward sub-layer, as shown in the left part of Fig.~\ref{fig2}. The self-attention sub-layer calculates the output representation of a token by attending to all the neighbors in the same layer, computing the correlation score between this token and all the neighbors,  and finally linearly combining all the representations of the neighbors and itself. The output of the $N$-th encoder layer is the source-side semantic representation ${\bm{h}}$. The decoder as shown in the right part in Fig.~\ref{fig2} also consists of $N$ identical layers. Each layer has three sub-layers. The first one is the masked self-attention mechanism that summarizes the partial prediction history. The second one is the encoder-decoder attention sub-layer determining the dynamic source-side contexts for current prediction and the third one is the feed-forward sub-layer. Residual connection and layer normalization are performed for each sub-layer in both of the encoder and decoder.

It is easy to notice that the attention mechanism is the key component. There are three kinds of  attention mechanisms, including encoder self-attention, decoder masked self-attention and encoder-decoder attention. They can be formalized into the same formula.

\begin{equation}
\text{Attention}({\bm{q}},{\bm{K}},{\bm{V}})=\text{softmax}{\bigg(}\frac{{\bm{q}}{\bm{K}}^T}{\sqrt{d_k}}{\bigg)}{\bm{V}}
\label{transformerattention}
\end{equation}
where ${\bm{q}}$, ${\bm{K}}$ and ${\bm{V}}$ stand for a query, the key list and the value list respectively. $d_k$ is the dimension of the key.

For the encoder self-attention, the queries, keys and values are from the same layer. For example, considering we calculate the output of the first layer in the encoder at the $j$-th position, let ${\bm{x}}_j$ be the sum vector of input token embedding and the positional embedding. The query is vector ${\bm{x}}_j$. The keys and values are the same and both are the embedding matrix ${\bm{x}}=[{\bm{x}}_0 \cdots {\bm{x}}_{J}]$. Then, multi-head attention is proposed to calculate attentions in different subspaces.

\begin{equation}
\begin{aligned} \label{MHAtt}
\mbox{MultiHead}(q,{\textbf{K}}, {\textbf{V}}) = \mbox{Concat}_i(head_i){\textbf{W}}_O  \\
head_i = \mbox{Attention}({\textbf{q}}{\textbf{W}}_Q^i, {\textbf{K}}{\textbf{W}}_K^i, {\textbf{V}}{\textbf{W}}_V^i)
\end{aligned}
\end{equation}
in which ${\textbf{W}}_Q^i,$, ${\textbf{W}}_K^i$, ${\textbf{W}}_V^i$ and ${\textbf{W}}_O$ denote projection parameter matrices .

Using Equation~\ref{MHAtt} followed by residential connection, layer normalization and a feed-forward network, we can get the representation of the second layer. After $N$ layers, we obtain the input contexts ${\bm{C}}=[{\bm{h}}_0, \cdots, {\bm{h}}_{J}]$.

The decoder masked self-attention is similar to that of encoder except that the query at the $i$-th position can only attend to positions before $i$, since the predictions after $i$-th position are not available in the auto-regressive left-to-right unidirectional inference.

\begin{equation}
{\bm{z}}_i=\text{Attention}({\bm{q}}_i,{\bm{K}}_{\le i},{\bm{V}}_{\le i})=\text{softmax}(\frac{{\bm{q}}_i{\bm{K}}^T_{\le i}}{\sqrt{d_k}}){\bm{V}}_{\le i}
\label{history-attention}
\end{equation}

The encoder-decoder attention mechanism is to calculate the source-side dynamic context which is responsible to predict the current target-language token. The query is the output of the masked self-attention sub-layer ${\bm{z}}_i$. The keys and values are the same encoder contexts $\bm{C}$. The residential connection, layer normalization and feed-forward sub-layer are then applied to yield the output of a whole layer. After $N$ such layers, we obtain the final hidden state ${\bm{z}}_i$. Softmax function is then employed to predict the output ${y}_i$, as shown in the upper right part of Fig.~\ref{fig2}.

\section{Key challenging research tasks}\label{section3}
Although Transformer has significantly advanced the development of neural machine translation, many challenges still remain to be addressed. Obviously, designing better NMT framework must be the most important challenge. However, since the innovation of Transformer, almost no more effective NMT architecture has been proposed. \cite{chen2018best} presented an alternative encoder-decoder framework RNMT+ which combines the merits of RNN-based and Transformer-based models to perform translation. \cite{wang2019learning,zhang2019improving} investigated how to design much deeper Transformer model and \cite{li2020neural} presented a Reformer model enabling rich interaction between encoder and decoder. \cite{wu2019pay} attempted to replace self-attention with dynamic convolutions. \cite{so2019evolved} proposed the evolved Transformer using neural architecture search. \cite{lu2019understanding} aimed to improve Transformer from the perspective of multi-particle dynamic system. Note that these models do not introduce big change on the NMT architecture. Pursuing to design novel and more effective NMT framework will be a long way to go.
In this section, we analyze and discuss the key challenges{\footnote{\cite{zhang2015deep,liu2018deep,koehn2017six} have also discussed various challenges.}} facing NMT from its formulation.

From the introduction in Sec.~\ref{subsection21}, NMT is formally defined as a sequence-to-sequence prediction task in which four assumptions are hidden in default. First, the input is a sentence rather than paragraphs and documents. Second, the output sequence is generated in a left-to-right autoregressive manner. Third, the NMT model is optimized over the bilingual training data which should include large-scale parallel sentences in order to learn good network parameters. Fourth, the processing objects of NMT are the pure texts (tokens, words and sentences) instead of speech and videos.  Accordingly, four key challenges can be summarized as follows:

\begin{itemize}
\item[1.] {\bf Document neural machine translation}. In NMT formulation, sentence is the basic input for modeling. However, some words in the sentence are ambiguous and the sense can only be disambiguated with the context of surrounding sentences or paragraphs. And when translating a document, we need to guarantee the same terms in different sentences lead to the same translation while performing translation sentence by sentence independently cannot achieve this goal. Moreover, many discourse phenomena such as coreference, omissions and coherence, cannot be handled in the absence of document-level information. Obviously, it is a big challenge how to take full advantage of contexts beyond sentences in neural machine translation.

\item[2.] {\bf Non-autoregressive decoding and bidirectional inference}. Left-to-right decoding token by token follows an autoregressive style which seems natural and is in line with human reading and writing. It is also easy for training and inference. However, it has several drawbacks. On one hand, the decoding efficiency is quite limited since the $i$-th translation token can be predicted only after all the previous $i-1$ predictions have been generated. On the other hand, predicting the $i$-th token can only access the previous history predictions while cannot utilize the future context information in autoregressive manner, leading to inferior translation quality. Thus, it is a challenge how to break the autoregressive inference constraint. Non-autoregressive decoding and bidirectional inference are two solutions from the perspectives of efficiency and quality respectively.

\item[3.] {\bf Low-resource translation}. There are thousands of human languages in the world and abundant bitexts are only available in a handful of language pairs such as English-German, English-French and English-Chinese. Even in the resource-rich language pair, the parallel data are unbalanced since most of the bitexts mainly exist in several domains (e.g. news and patents). That is to say, the lack of parallel training corpus is very common in most languages and domains. It is well-known that neural network parameters can be well optimized on highly repeated events (frequent word/phrase translation pairs in the training data for NMT) and the standard NMT model will be poorly learned on low-resource language pairs. As a result, how to make full use of the parallel data in other languages (pivot-based translation and multilingual translation) and how to take full advantage of non-parallel data (semi-supervised translation and unsupervised translation) are two challenges facing NMT.

\item[4.] {\bf Multimodal neural machine translation}. Intuitively, human language is not only about texts and understanding the meaning of a language may need the help of other modalities such as speech, image and videos. Concerning the well-known example that determines the meaning of the word \emph{bank} when translating the sentence "he went to the bank", it will be correctly translated if we are given an image in which a man is approaching a river. Furthermore, in many scenarios, we are required to translate a speech or a video. For example, simultaneous speech translation is more and more demanding in various conferences or international live events. Therefore, how to perform multimodal translation under the encoder-decoder architecture is a big challenge of NMT. How to make full use of different modalities in multimodal translation and how to balance the quality and latency in simultaneous speech translation are two specific challenges.

\end{itemize}

In the following sections, we briefly introduce the recent progress for each challenge.

\section{Document-level neural machine translation}\label{section4}

\begin{figure*}[t]
	\centering
	\includegraphics[scale=.6]{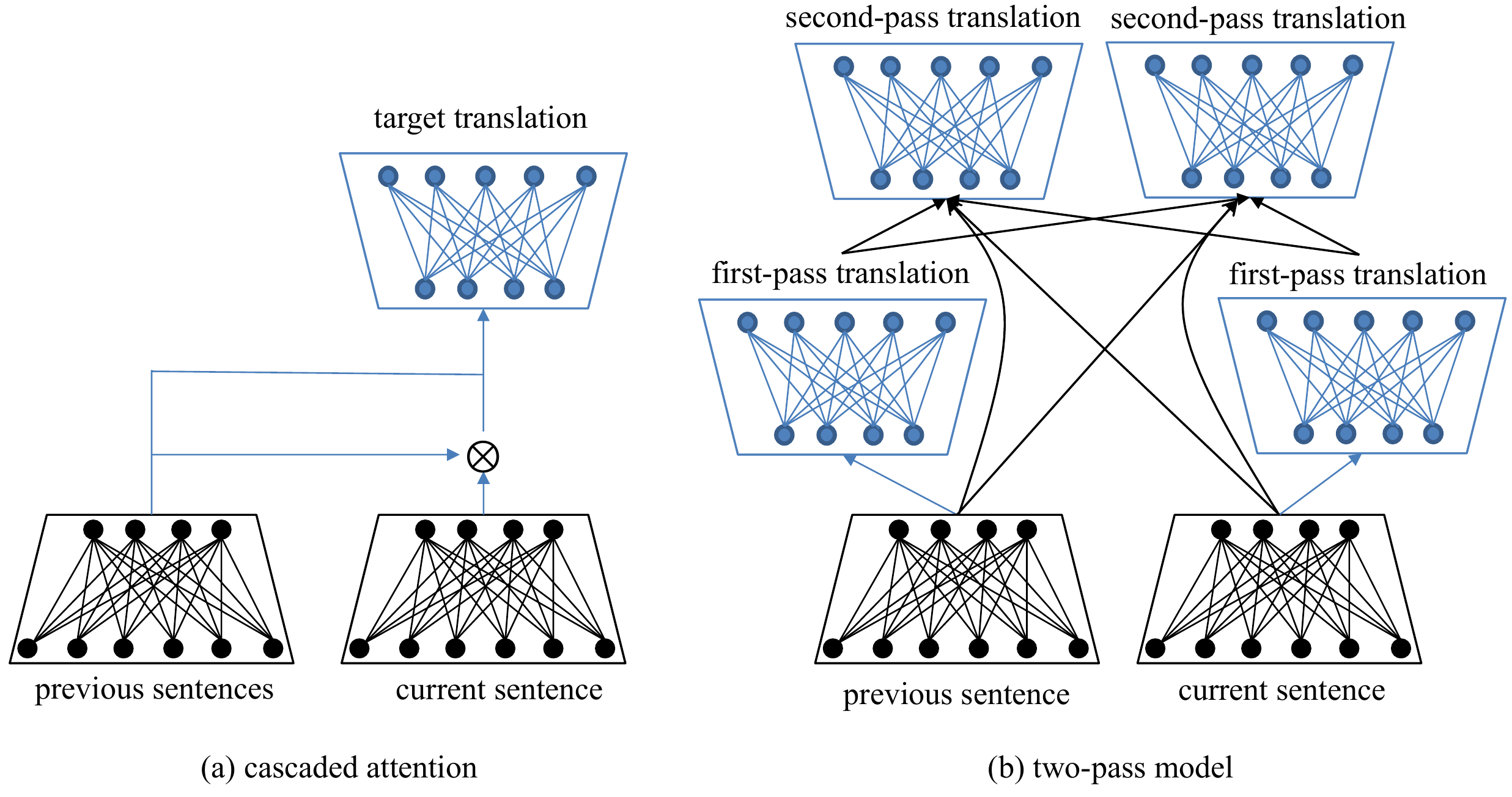}
	\caption{illustration of two docNMT models. The left part shows the cascaded attention model proposed by \cite{zhang2018improving} in which the previous source sentences are first leveraged to enhance the representation of current source sentence and then used again in the decoder. The right part illustrates the two-pass docNMT model proposed by \cite{xiong2019modeling} in which sentence-level NMT first generates preliminary translation for each sentence and then the first-pass translations together with the source-side sentences are employed to generate the final translation results.}
	\label{fig7}
\end{figure*}

As we discussed in Sec.~\ref{section3} that performing translation sentence by sentence independently would introduce several risks. An ambiguous word may not be correctly translated without the necessary information in the surrounding contextual sentences. A same term in different sentences in the same document may result in inconsistent translations. Furthermore, many discourse phenomena, such as coreference, omissions and cross-sentence relations, cannot be well handled. In a word, sentence-level translation will harm the coherence and cohesion of the translated documents if we ignore the discourse connections and relations between sentences.

In general, document-level machine translation (docMT) aims at exploiting the useful document-level information (multiple sentences around the current sentence or the whole document) to improve the translation quality of the current sentence as well as the coherence and cohesion of the translated document. docMT has already been extensively studied in the era of statistical machine translation (SMT), in which most researchers mainly propose explicit models to address some specific discourse phenomena, such as lexical cohesion and consistency \cite{gong2011cache,xiao2011document,xiong2013lexical}, coherence \cite{born2017using} and coreference \cite{rios2017co}. Due to complicate integration of multiple components in SMT, these methods modeling discourse phenomenon do not lead to promising improvements.

The NMT model dealing with semantics and translation in the distributed vector space facilitates the use of wider and deep document-level information under the encoder-decoder framework. It does not need to explicitly model specific discourse phenomenon as that in SMT. According to the types of used document information, document-level neural machine translation (docNMT) can roughly fall into three categories: dynamic translation memory \cite{kuang2018modeling,tu2018learning}, surrounding sentences \cite{jean2017does,wang2017exploiting,voita2018context,zhang2018improving,miculicich2018document,xiong2019modeling,yang2019enhancing} and the whole document \cite{maruf2018document,maruf2019selective,tan2019hierarchical}.

\cite{tu2018learning} presented a dynamic cache-like memory to maintain the hidden representations of previously translated words. The memory contains a fixed number of cells and each cell is a triple $(\bm{c}_t, \bm{s}_t, y_t)$ where $y_t$ is the prediction at the $t$-th step, $\bm{c}_t$ is the source-side context representation calculated by the attention model and $\bm{s}_t$ is the corresponding decoder state. During inference, when predicting the $i$-th prediction for a test sentence, $\bm{c}_i$ is first obtained through attention model and the probability $p(\bm{c}_t|\bm{c}_i)$ is computed based on their similarity. Then memory context representation $\bm{m}_i$ is calculated by linearly combining all the values $\bm{s}_t$ with $p(\bm{c}_t|\bm{c}_i)$. This cache-like memory can encourage the words in similar contexts to share similar translations so that cohesion can be enhanced to some extent.

The biggest difference between the use of whole document and surrounding sentences lies in the number of sentences employed as the context. This article mainly introduces the methods exploiting surrounding sentences for docNMT. Relevant experiments further show that subsequent sentences on the right contribute little to the translation quality of the current sentence. Thus, most of the recent work aim at fully exploring the previous sentences to enhance docNMT. These methods can be divided into two categories. One just utilizes the previous source-side sentences \cite{tiedemann2017neural,jean2017does,wang2017exploiting,zhang2018improving}. The other uses the previous source sentences as well as their target translations \cite{miculicich2018document,xiong2019modeling}.

If only previous source-side sentences are leveraged, the previous sentences can be concatenated with the current sentence as the input to the NMT model \cite{tiedemann2017neural} or could be encoded into a summarized source-side context with a hierarchical neural network \cite{wang2017exploiting}. \cite{zhang2018improving} presented a cascaded attention model to make full use of the previous source sentences. As shown in the left part of Fig.~\ref{fig7}, previous sentence is first encoded as the document-level context representation. When encoding the current sentence, each word will attend to the document-level context and obtain a context-enhanced source representation. During the calculation of cross-language attention in the decoder,  the current source sentence together with the document-level context are both leveraged to predict the target word. The probability of translation sentence given the current sentence and the previous context sentences is formulated as follows:

\begin{equation}
P(\bm{y}|\bm{x}, \text{doc}_x; \theta)=\prod_{i=0}^{I}P(y_i|\bm{y}_{<i}, \bm{x}, \text{doc}_x; \theta)
\end{equation}
where $\text{doc}_x$ denotes the source-side document-level context, namely previous sentences.

If both of previous source sentences and their translations are employed, two-pass decoding is more suitable for the docNMT model \cite{xiong2019modeling}. As illustrated in the right part of Fig.~\ref{fig7}, the sentence-level NMT model can generate preliminary translations for each sentence in the first-pass decoding. Then, the second-pass model will produce final translations with the help of source sentences and their preliminary translation results. The probability of the target sentence in the second pass can be written by:

\begin{equation}
P(\bm{y}|\bm{x}, \text{doc}_x, \text{doc}_y; \theta)=\prod_{i=0}^{I}P(y_i|\bm{y}_{<i}, \bm{x}, \text{doc}_x, \text{doc}_y; \theta)
\end{equation}
in which $\text{doc}_y$ denotes the first-pass translations of $\text{doc}_x$.

Since most methods for docNMT are designed to boost the overall translation quality (e.g. BLEU score), it still remains a big problem whether these methods indeed well handle the discourse phenomena. To address this issue, \cite{bawden2018evaluating} conducted an empirical investigation of the docNMT model on the performance of processing various discourse phenomena, such as coreference, cohesion and coherence. Their findings indicate that multi-encoder model exploring only the source-side previous sentences performs poorly in handling the discourse phenomena while exploiting both source sentences and target translations leads to the best performance. Accordingly, \cite{voita2019good,voita2019context} recently focused on designing better document-level NMT to improve on specific discourse phenomena such as deixis, ellipsis and lexical cohesion for English-Russian translation.

\begin{figure*}[t]
	\centering
	\includegraphics[scale=.6]{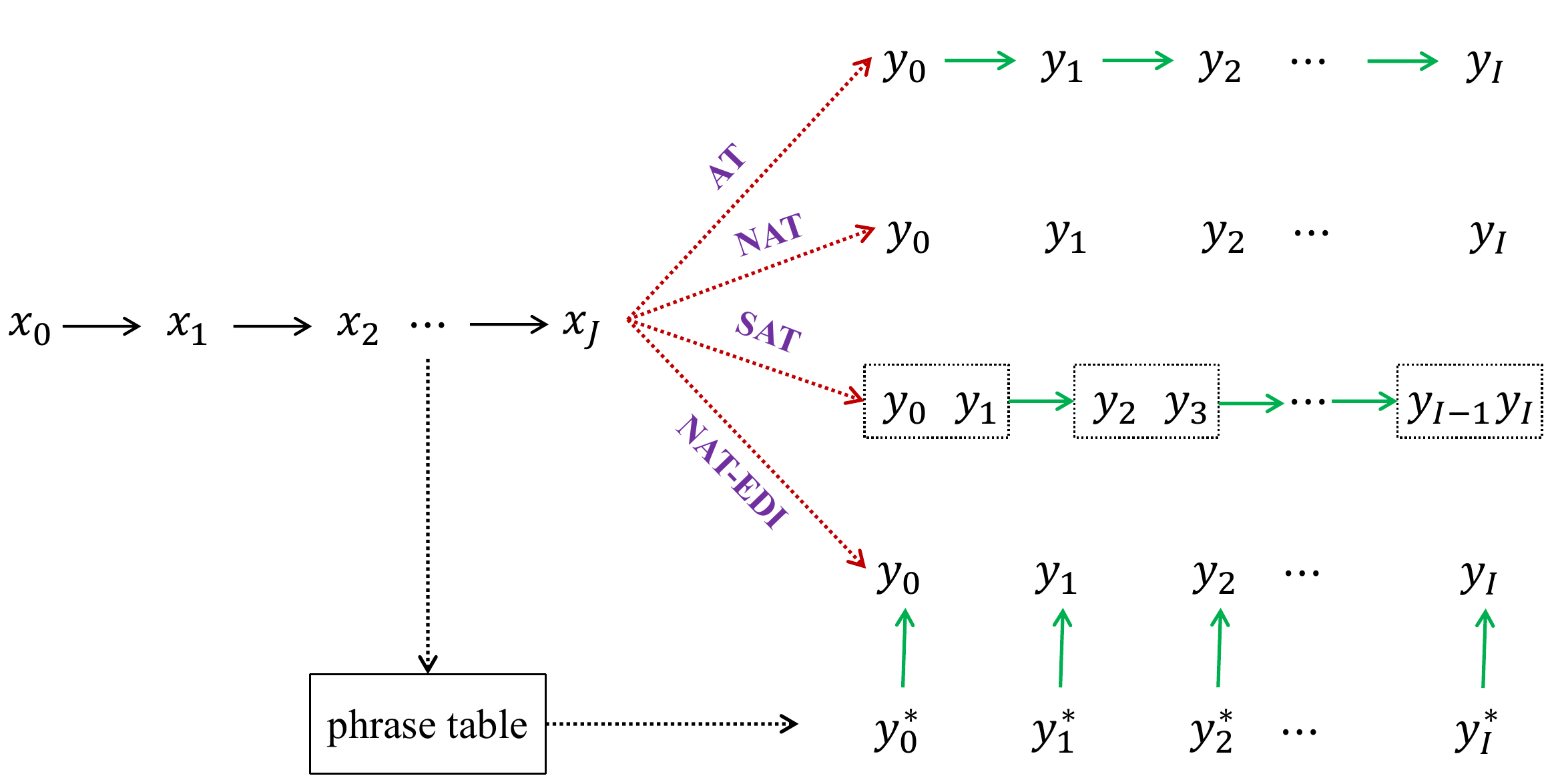}
	\caption{illustration of autoregressive NMT model and various non-autoregressive NMT models. {\bf{AT}} denotes the conventional autoregressive NMT paradigm in which the $i$-th prediction can fully utilize the partial translation of $i-1$ words. {\bf{NAT}} indicates the non-autoregreesive NMT model that generates all the target words simultaneously. {\bf{SAT}} is a variant of {\bf{NAT}} which produces an ngram each time. {\bf{NAT-EDI}} denotes the non-autoregressive NMT model with enhanced decoder input which is generated by retrieving the phrase table.}
	\label{fig8}
\end{figure*}

\section{Non-autoregressive decoding and bidirectional inference}\label{section5}
Most NMT models follow the autoregressive generation style which produces output word by word from left to right. Just as Sec.~\ref{section3} discussed, this paradigm has to wait for $i-1$ time steps before starting to predict the $i$-th target word. Furthermore, left-to-right autoregressive decoding cannot exploit the target-side future context (future predictions after $i$-th word). Recently, many research work attempt to break this decoding paradigm. Non-autoregressive Transformer (NAT) \cite{gu2018non} is proposed to remarkably lower down the latency by emitting all of the target words at the same time and bidirectional inference \cite{zhang2018asynchronous,zhou2019synchronous} is introduced to improve the translation quality by making full use of both history and future contexts.

\subsection{Non-autoregressive decoding}\label{section51}
Non-autoregressive Transformer (NAT) aims at producing an entire target output in parallel. Different from the autoregressive Transformer model (AT) which terminates decoding when emitting an end-of-sentence token $\langle /s \rangle$, NAT has to know how many target words should be generated before parallel decoding. Accordingly, NAT calculates the conditional probability of a translation $\bm{y}$ given the source sentence $\bm{x}$ as follows:

\begin{equation}
P_{NAT}(\bm{y}|\bm{x};\theta)=P_L(I|\bm{x};\theta)\cdot \prod_{i=0}^{I}P(y_i|\bm{x};\theta)
\end{equation}

To determine the output length, \cite{gu2018non} proposed to use the fertility model which predicts the number of target words that should be translated for each source word. We can perform word alignment on the bilingual training data to obtain the gold fertilities for each sentence pair. Then, the fertility model can be trained together with the translation model. For each source word $x_j$, suppose the predicted fertility is $\Phi(x_j)$. The output length will be $I=\sum_{j=0}^{J}\Phi(x_j)$.

Another issue remains that AT let the previous generated output $y_{i-1}$ be the input at the next time step to predict the $i$-th target word but NAT has no such input in the decoder network. \cite{gu2018non} found that translation quality is particularly poor if omitting the decoder input in NAT. To address this, they resort to the fertility model again and copy each source word as many times as its fertility $\Phi(x_j)$ into the decoder input. The empirical experiments show that NAT can dramatically boost the decoding efficiency by $15\times$ speedup compared to AT. However, NAT severely suffers from accuracy degradation.

The low translation quality may be due to at least two critical issues of NAT. First, there is no dependency between target words although word dependency is ubiquitous in natural language generation. Second, the decoder inputs are the copied source words which lie in different semantic space with target words. Recently, to address the shortcomings of the original NAT model, several methods are proposed to improve the translation quality of NAT while maintaining its efficiency \cite{lee2018deterministic,wang2018semi,guo2019non,shao2019retrieving,wang2019non,wei2019imitation}. 

\cite{wang2018semi} proposed a semi-autoregressive Transformer model (SAT) to combine the merits of both AT and NAT. SAT keeps the autoregressive property in global but performs NAT in local. Just as shown in Fig.~\ref{fig8}, SAT generates $K$ successive target words at each time step in parallel. If $K=1$, SAT will be exactly AT. It will become NAT if $K=I$. By choosing an appropriate $K$, dependency relation between fragments are well modeled and the translation quality can be much improved with some loss of efficiency.

\begin{figure*}[t]
	\centering
	\includegraphics[scale=.5]{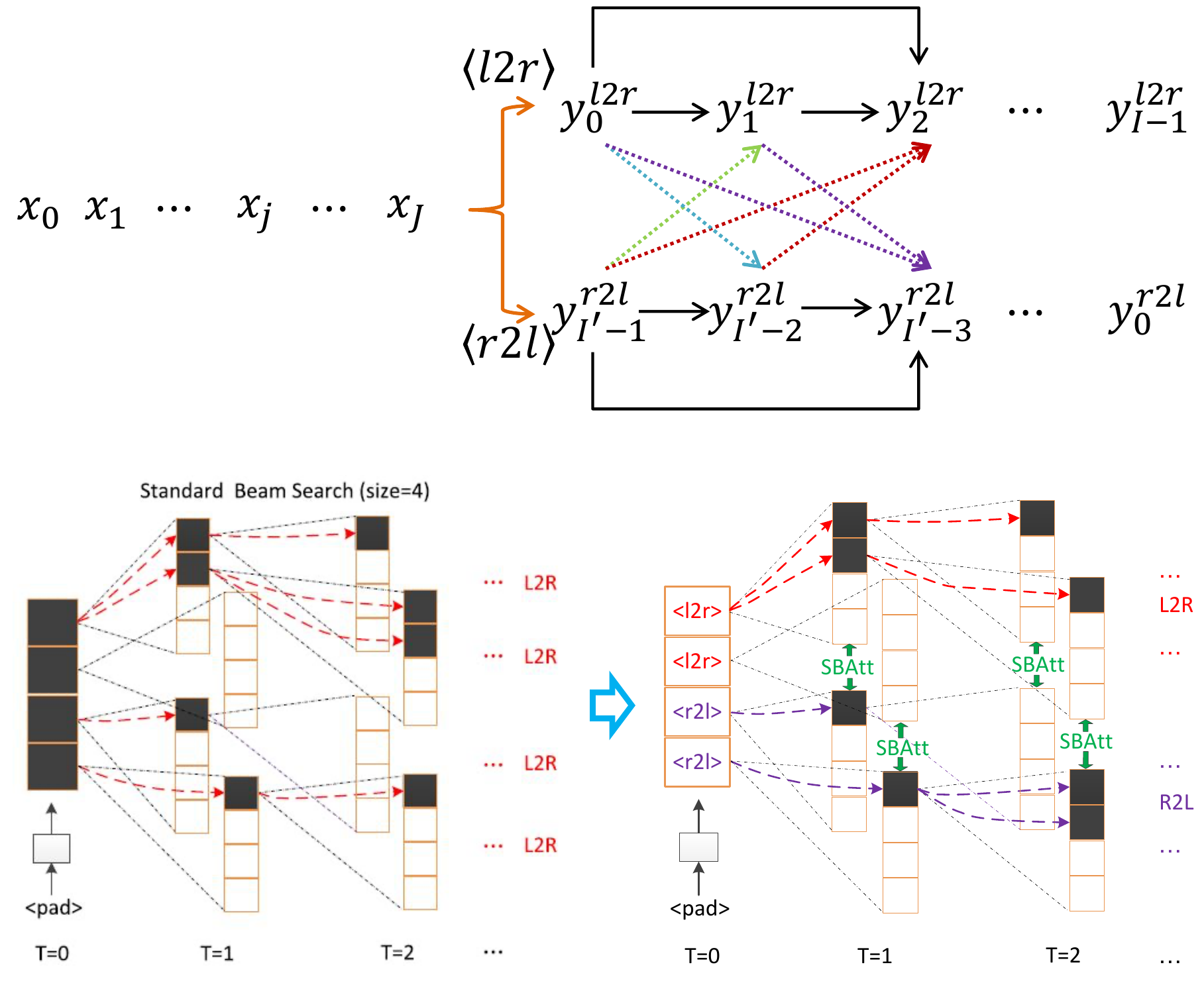}
	\caption{illustration of the synchronous bidirectional inference model. The top demonstrates how the bidirectional contexts can be leveraged during inference. The bottom compares the beam search algorithm between the conventional NMT and the synchronous bidirectional NMT.}
	\label{fig9}
\end{figure*}

To mimic the decoder input in the AT model, \cite{guo2019non} introduced a simple but effective method that employs a phrase table which is the core component in SMT to convert source words into target words. Specifically, they first greedily segment the source sentence into phrases with maximum match algorithm. Suppose the longest phrase in the phrase table contains $K$ words. $\bm{x}_{0:K-1}$ is a phrase if it matches an entry in the phrase table. Otherwise they iteratively check $\bm{x}_{0:K-2}$, $\bm{x}_{0:K-3}$ and so on. If $\bm{x}_{0:h}$ is a phrase, then they start to check $\bm{x}_{h+1:h+K}$. After segmentation, each source phrase is mapped into target translations which are concatenated together as the new decoder input, as shown in Fig.~\ref{fig8}. Due to proper modeling of the decoder input with a highly efficient strategy, translation quality is substantially improved while the decoding speed is even faster than baseline NAT.

\subsection{Bidirectional inference}\label{section52}

From the viewpoint of improving translation quality, autoregressive model can be enhanced by exploring the future context on the right. In addition to predicting and estimating the future contexts with various models \cite{zheng2018modeling,zheng2019dynamic,zhang2019future}, researchers find that left-to-right (L2R) and right-to-left (R2L) autoregressive models can generate complementary translations \cite{liu2016agreementa,hoang2017decoding,zhang2018asynchronous,zhou2019synchronous}. For example, in Chinese-to-English translation, experiments show that L2R can generate better prefix while R2L is good at producing suffix. Intuitively, it is a promising direction to combine the merits of bidirectional inferences and fully exploit both history and future contexts on the target side.

To this end, many researchers resort to exploring bidirectional decoding to take advantages of both L2R and R2L inferences. These methods are mainly fall into four categories: 1, enhancing agreement between L2R and R2L predictions \cite{liu2016agreementa,zhang2019regularizing}; 2, reranking with bidirectional decoding \cite{liu2016agreementa,sennrich2016edinburgh,sennrich2017university}; 3, asynchronous bidirectional decoding \cite{zhang2018asynchronous,su2019exploiting} and 4, synchronous bidirectional decoding \cite{zhou2019synchronous,zhou2019sequence,zhang2020synchronous}.

Ideally, L2R decoding should generate the same translation as R2L decoding. Under this reasonable assumption, \cite{liu2016agreementa,zhang2019regularizing} introduced agreement constraint or regularization between L2R and R2L predictions during training. Then, L2R inference can be improved.

The reranking algorithm is widely used in machine translation, and the R2L model can provide an estimation score for the quality of L2R translation from another parameter space \cite{liu2016agreementa,sennrich2016edinburgh,sennrich2017university}. Specifically, L2R first generates a $n$-best list of translations. The R2L model is then leveraged to force decode each translation leading to a new score. Finally, the best translation is selected according to the new scores.

\cite{zhang2018asynchronous,su2019exploiting} proposed an asynchronous bidirectional decoding model (ASBD) which first obtains the R2L outputs and optimizes the L2R inference model based on both of the source input and the R2L outputs. Specifically, \cite{zhang2018asynchronous} first trained a R2L model with the bilingual training data. Then, the optimized R2L decoder translates the source input of each sentence pair and produces the outputs (hidden states) which serve as the additional context for L2R prediction when optimizing the L2R inference model. Due to explicit use of right-side future contexts, the ASBD model significantly improves the translation quality. But these approaches still suffer from two issues. On one hand, they have to train two separate NMT models for L2R and R2L inferences respectively. And the two-pass decoding strategy makes the latency much increased. On the other hand, the two models cannot interact with each other during inference, which limits the potential of performance improvement.

\cite{zhou2019synchronous} proposed a synchronous bidirectional decoding model (SBD) that produces outputs using both L2R and R2L decoding simultaneously and interactively. Specifically, a new synchronous attention model is proposed to conduct interaction between L2R and R2L inferences. The top part in Fig.~\ref{fig9} gives a simple illustration of the proposed synchronous bidirectional inference model. The dotted arrows in color on the target side is the core of the SBD model. L2R and R2L inferences interact with each other in an implicit way illustrated by the dotted arrows. All the arrows indicate the information passing flow. Solid arrows show the conventional history context dependence while dotted arrows introduce the future context dependence on the other inference direction.  For example, besides the past predictions ($y_0^{l2r}$, $y_1^{l2r}$), L2R inference can also utilize the future contexts ($y_0^{r2l}$, $y_1^{r2l}$)  generated by the R2L inference when predicting $y_2^{l2r}$. The conditional probability of the translation can be written as follows:

\begin{equation}
P(\bm{y}|\bm{x}) =
\begin{cases}
\prod_{i=0}^{I} P(\overrightarrow{y_i}|\overrightarrow{y}_0\cdots \overrightarrow{y}_{i-1}, x, \overleftarrow{y}_0\cdots \overleftarrow{y}_{i-1}) & \text{if L2R} \\
\prod_{i=0}^{I'-1} P(\overleftarrow{y}_i|\overleftarrow{y}_0\cdots \overleftarrow{y}_{i-1}, x, \overrightarrow{y}_0\cdots \overrightarrow{y}_{i-1}) & \text{if R2L}
\end{cases}
\label{bidecoposition}
\end{equation}

To accommodate L2R and R2L inferences at the same time, they introduced a novel beam search algorithm. As shown in the bottom right of Fig.~\ref{fig9}, at each timestep during decoding, each half beam maintains the hypotheses from L2R and R2L decoding respectively and each hypothesis is generated by leveraging already predicted outputs from both directions. At last, the final translation is chosen from L2R and R2L results according to their translation probability. Thanks to appropriate rich interaction, the SBD model substantially boosts the translation quality while the decoding speed is only 10\% slowed down.

\cite{zhou2019sequence} further noticed that L2R and R2L are not necessary to produce the entire translation sentence. They let L2R generate the left half translation and make R2L produce the right half, and then two halves are concatenated to form the final translation. Using proper training algorithms, they demonstrated through extensive experiments that both translation quality and decoding efficiency can be significantly improved compared to the baseline Transformer model.

\section{Low-resource Translation}\label{section6}
Most NMT models assume that enough bilingual training data is available, which is the rare case in real life. For a low-resource language pair, a natural question may arise that what kind of knowledge can be transferred to build a relatively good NMT system. This section will discuss three kinds of methods. One attempts to share translation knowledge from other resource-rich language pairs, in which pivot translation and multilingual translation are the two key techniques. Pivot translation assumes that for the low-resource pair $A$ and $B$, there is a language $C$ that has rich bitexts with $A$ and $B$ respectively \cite{wu2007pivot,cheng2017joint}. This section mainly discusses the technique of multilingual translation in the first category. The second kind of methods resort to semi-supervised approach which takes full advantages of limited bilingual training data and abundant monolingual data. The last one leverages unsupervised algorithm that requires monolingual data only.

\subsection{Multilingual neural machine translation}\label{subsection61}

Let us first have a quick recap about the NMT model based on encoder-decoder framework. The encoder is responsible for mapping the source language sentence into distributed semantic representations. The decoder is to convert the source-side distributed semantic representations into target language sentence. Apparently, the encoder and the decoder (excluding the cross-language attention component) are just single-language dependent.  Intuitively, the same source language in different translation systems (e.g. Chinese-to-English, Chinese-to-Hindi) can share the same encoder and the same target language can share the same decoder (e.g. Chinese-to-English and Hindi-to-English). Multilingual neural machine translation is a framework that aims at building a unified NMT model capable of translating multiple languages through parameter sharing and knowledge transferring.

\cite{dong2015multi} is the first to design a multi-task learning method which shares the same encoder for one-to-many translation (one source language to multiple target languages). \cite{zoph2016multi} proposed to share the decoder for many-to-one translation (many source languages to one target language). \cite{firat2016multi,firat2017multi} proposed to share attention mechanism for many-to-many translation (many source languages to many target languages). Despite performance improved for low-resource languages, these methods are required to design a specific encoder or decoder for each language, hinders their scalability in dealing with many languages.

\cite{johnson2017google} goes a step further and let all source languages share the same encoder and all the target languages share the same decoder. They have successfully trained a single encoder-decoder NMT model for multilingual translation. The biggest issue is that the decoder is unaware of which target language should be translated to at the test phase. To this end, \cite{johnson2017google} introduced a simple strategy and added a special token indicating target language (e.g {\emph{2en}} and {\emph{2zh}}) at the beginning of the source sentence. By doing this, low-resource languages have the biggest chance to share translation knowledge from other resource-rich languages. It also enables zero-shot translation as long as the two languages are employed as source and target in the multilingual NMT model. In addition, this unified multilingual NMT is very scalable and could translate all the languages in one model ideally. However, experiments find that the output is sometimes mixed of multiple languages even using a translation direction indicator. Furthermore, this paradigm enforces different source/target languages to share the same semantic space, without considering the structural divergency among different languages. The consequence is that the single model based multilingual NMT yields inferior translation performance compared to individually trained bilingual counterparts. Most of recent research work mainly focus on designing better models to well balance the language-independent parameter sharing and the language-sensitive module design.

\cite{blackwood2018multilingual} augmented the attention mechanism in decoder with language-specific signals. \cite{wang2018three} proposed to use language-sensitive positions and language-dependent hidden presentations for one-to-many translation. \cite{platanios2018contextual} designed an algorithm to generate language-specific parameters. \cite{tan2019multilingual} designed a language clustering method and forced languages in the same cluster to share the parameters in the same semantic space. \cite{wang2019synchronously} attempted to generate two languages simultaneously and interactively by sharing encoder parameters. \cite{wang2019compact} proposed a compact and language-sensitive multilingual translation model which attempts to share most of the parameters while maintaining the language discrimination.

\begin{figure}[H]
	\centering
	\includegraphics[scale=.5]{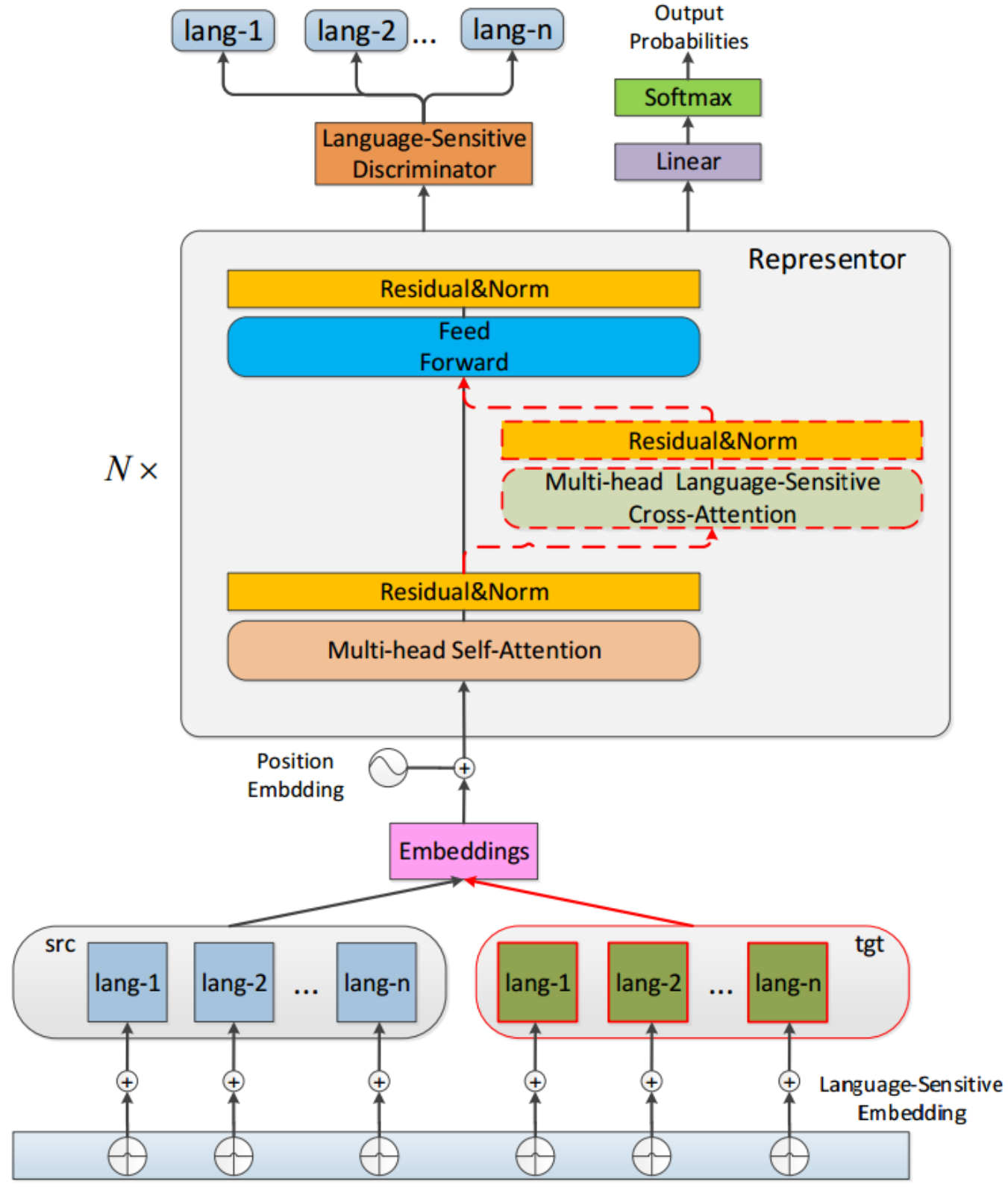}
	\caption{illustration of a compact and language-sensitive multilingual NMT model. The compactness is ensured by sharing parameters between encoder and decoder, denoted as representor. Language-sensitive capacity is realized by three components: language-sensitive embedding (bottom), language-sensitive cross-attention (middle) and language discriminator (top).}
	\label{fig10}
\end{figure}

As shown in Fig.~\ref{fig10}, \cite{wang2019compact} designed four novel modules in the Transformer framework compared to single-model based multilingual NMT. First, they introduced a representor to replace both encoder and decoder by sharing weight parameters of the self-attention block, feed-forward and normalization blocks (middle part in Fig.~\ref{fig10}). It makes the multilingual NMT model as compact as possible and maximizes the knowledge sharing among different languages. The objective function over $L$ language pairs becomes:

\begin{equation}
\mathcal{L}(\theta) = \sum_{l=1}^{L} \sum_{m=1}^{M_l} {\text{log}}P({\bm y}_l^{(m)}|{\bm x}_l^{(m)}; \theta_{rep}, \theta_{att})
\end{equation}
where $\theta_{rep}$ and $\theta_{att}$ denote parameters of representor and attention mechanism respectively.

However, the representor further reduces the ability to discriminate different languages. To address this, they introduced three language-sensitive modules. 

1. Language-sensitive embedding (bottom part in Fig.~\ref{fig10}): they compared four categories of embedding sharing patterns, namely language-based pattern (different languages have separate input embeddings), direction-based patter (languages in source side and target side have different input embeddings), representor-based pattern (shared input embeddings for all languages) and three-way weight tying pattern proposed by \cite{press2017using}, in which the output embedding of the target side is also shared besides representor-based sharing.

2. Language-sensitive attention (middle part in Fig.~\ref{fig10}): this mechanism allows the model to select the cross-lingual attention parameters according to specific translation tasks dynamically.

3. Language-sensitive discriminator (top part in Fig.~\ref{fig10}): for this module, they employed a neural model $f_{dis}$ on the top layer of reprensentor $\bm{h}_{top}^{rep}$, and this model outputs a language judgment score $P_{lang}$.

\begin{equation}
P_{lang}=\text{softmax}(W_{dis}\times f_{dis}(\bm{h}_{top}^{rep})+b_{dis})
\end{equation}

Combining the above four ideas together, they showed through extensive experiments that the new method significantly improves multilingual NMT on one-to-many, many-to-many and zero-shot scenarios, outperforming bilingual counterparts in most cases. It indicates that low-resource language translation can greatly benefit from this kind of multilingual NMT, and so do zero-resource language translations.

\begin{figure*}[t]
	\centering
	\includegraphics[scale=.6]{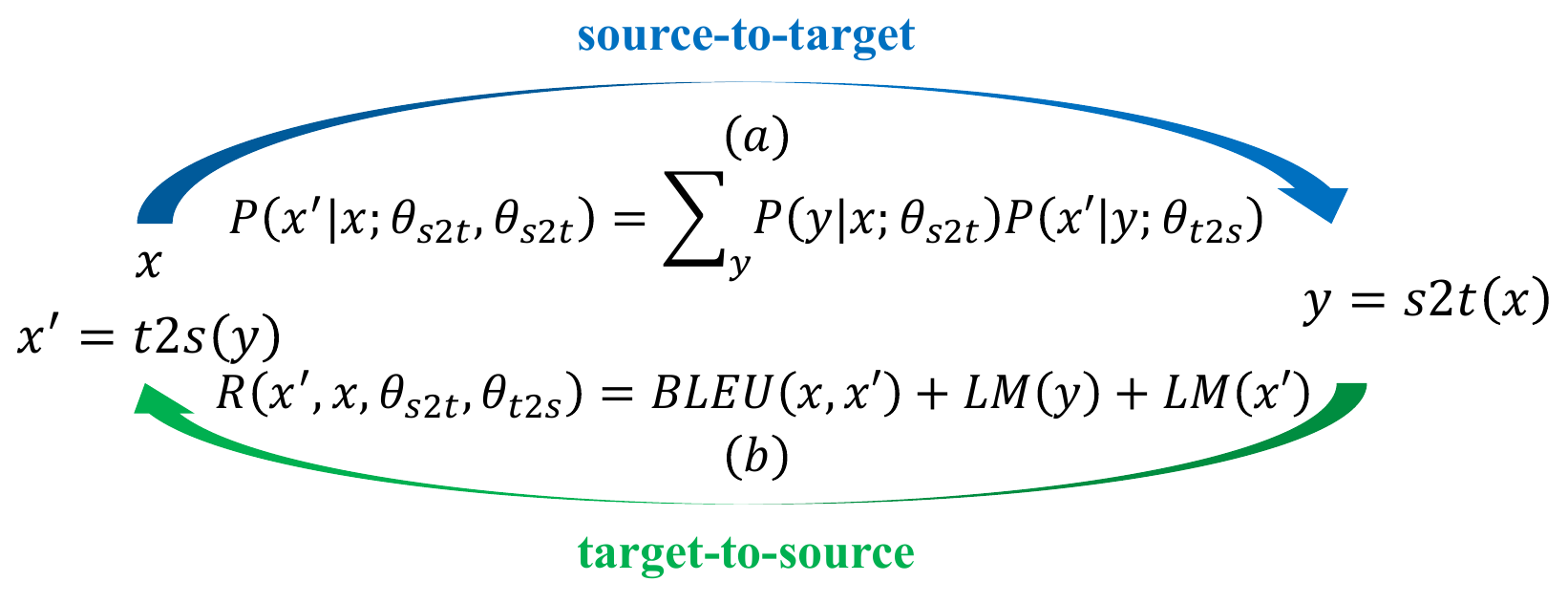}
	\caption{illustration of two methods exploring monolingual data. If the parameters are trained to maximize the objective function of (a), it is the auto-encoder based method. If using reward as (b) shows, it is the dual learning method. Note that this figure only demonstrates the usage of source-side monolingual data for simplicity. The use of target-side monolingual data is symmetric.}
	\label{fig11}
\end{figure*}

\subsection{Semi-supervised neural machine translation}\label{subsection62}

Semi-supervised neural machine translation is a paradigm which aims at building a good NMT system with limited bilingual training data $\mathcal{D}=\{\bm{x}^{(m)},\bm{y}^{(m)}\}_{m=1}^M$ plus massive source monolingual data $\mathcal{D}_{x}=\{\bm{x}^{(l_x)}\}_{l_x=1}^{L_x}$ or target monolingual data $\mathcal{D}_{y}=\{\bm{y}^{(l_y)}\}_{l_y=1}^{L_y}$ or both.

Monolingual data plays a very important role in SMT where the target-side monolingual corpus is leveraged to train a language model (LM) to measure the fluency of the translation candidates during decoding \cite{koehn2003statistical,chiang2005hierarchical,koehn2007moses}. Using monolingual data as a language model in NMT is not trivial since it needs to modify the architecture of the NMT model. \cite{gulcehre2015using,gulcehre2017integrating} integrated NMT with LM by combining hidden states of both models, making the model much complicated.

As for leveraging the target-side monolingual data, back-translation (BT) proposed by \cite{sennrich2016improving} may be one of the best solutions up to now. BT is easy and simple to use since it is model agnostic to the NMT framework \cite{hoang2018iterative,edunov2018understanding}. It only requires to train a target-to-source translation system to translate the target-side monolingual sentences back into source language. The source translation and its corresponding target sentence are paired as pseudo bitexts which combined together with original bilingual training data to train the source-to-target NMT system. It has been proved to be particularly useful in low-resource translation \cite{karakanta2018neural}. \cite{edunov2018understanding} conducted a deep analysis to understand BT and investigate various methods for synthetic source sentence generation. \cite{wang2019improving} proposed to measure the confidence level of synthetic bilingual sentences so as to filter the noise.

In order to utilize the source-side monolingual data, \cite{zhang2016exploiting} proposed two methods: forward translation and multi-task learning. Forward translation is similar to BT, and the multi-task learning method performs source-to-target translation task and source sentence reordering task by sharing the same encoder.

Many researchers resort to use both side monolingual data in NMT at the same time \cite{cheng2016semi,he2016dual,zhang2018joint,zheng2020mirror}. We summarize two methods in Fig.~\ref{fig11}: the auto-encoder based semi-supervised learning method \cite{cheng2016semi} and the dual learning method \cite{he2016dual}. For a source-side monolingual sentence $x$, \cite{cheng2016semi} employed source-to-target translation as encoder to generate latent variable $y$ and leverage target-to-source translation as decoder to reconstruct the input leading to $x'$. They optimize the parameters by maximizing the reconstruction probability as shown in Fig.~\ref{fig11}(a). The target-side monolingual data is used in a symmetric way. Fig.~\ref{fig11}(b) shows the objective function for the dual learning method. \cite{he2016dual} treated source-to-target translation as the primal task and target-to-source translation as the dual task. Agent $A$ sends through the primal task a translation of the source monolingual sentence to the agent $B$. $B$ is responsible to estimate the quality of the translation with a language model and the dual task. The rewards including the similarity between the input $x$ and reconstructed one $x'$, and two language model scores $LM(y)$, $LM(x')$, are employed to optimize the network parameters of both source-to-target and target-to-source NMT models. Similarly, the target-side monolingual data is used in a symmetric way in dual learning.

\cite{zhang2018joint} introduced an iterative back-translation algorithm to exploit both source and target monolingual data with an EM optimization method. \cite{zheng2020mirror} proposed a mirror-generative NMT model, that explores the monolingual data by unifying the source-to-target NMT model, the target-to-source NMT model, and two language models. They showed better performance can be achieved compared to back-translation, iterative back-translation and dual learning.

\subsection{Unsupervised neural machine translation}\label{subsection63}

\begin{figure*}[t]
	\centering
	\includegraphics[scale=.6]{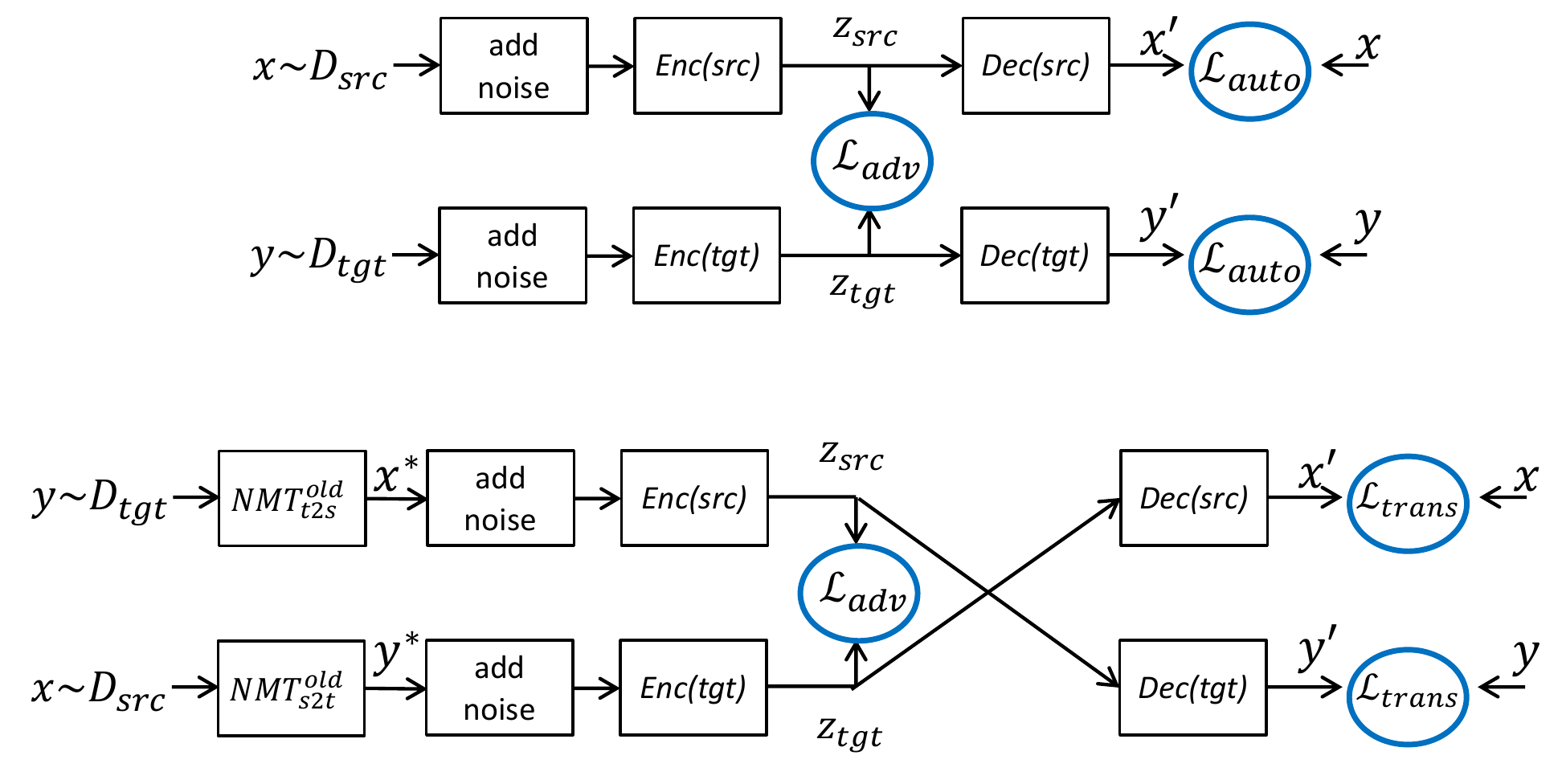}
	\caption{architecture of the unsupervised NMT model. The top shows denosing auto-encoder that aims at reconstructing the same language input. The bottom demonstrates back-translation which attempt to reconstruct the input in the other language using back-translation (target-to-source) and forward translation (source-to-target) The auto-encoder loss $\mathcal{L}_{auto}$, the translation loss $\mathcal{L}_{trans}$ and the language adversarial loss $\mathcal{L}_{adv}$ are used together to optimize the dual NMT models.}
	\label{fig12}
\end{figure*}

Unsupervised neural machine translation addresses a very challenging scenario in which we are required to build a NMT model using only 
massive source-side monolingual data $\mathcal{D}_{x}=\{\bm{x}^{(l_x)}\}_{l_x=1}^{L_x}$ and target-side monolingual data $\mathcal{D}_{y}=\{\bm{y}^{(l_y)}\}_{l_y=1}^{L_y}$.

Unsupervised  machine translation can date back to the era of SMT, in which decipherment approach is employed to learn word translations from monolingual data \cite{ravi2011deciphering,dou2012large,nuhn2012deciphering} or bilingual phrase pairs can be extracted and their probabilities can be estimated from monolingual data \cite{klementiev2012toward,zhang2013learning}.

Since \cite{mikolov2013exploiting} found that word embeddings from two languages can be mapped using some seed translation pairs, bilingual word embedding learning or bilingual lexicon induction has attracted more and more attention \cite{faruqui2014improving,zhang2017adversarial,zhang2017earth,artetxe2017learning,conneau2018word,cao2018point}. \cite{artetxe2017learning} and \cite{conneau2018word} applied linear embedding mapping and adversarial training to learn word pair matching in the distribution level and achieve promising accuracy for similar languages.

Bilingual lexicon induction greatly motivates the study of unsupervised NMT on sentence level. And two techniques of denoising auto-encoder and back-translation make it possible for unsupervised NMT. The key idea is to find a common latent space between the two languages. \cite{artetxe2018unsupervised} and \cite{lample2018unsupervised} both optimized dual tasks of source-to-target and target-to-source translation. \cite{artetxe2018unsupervised} employed shared encoder to force two languages into a same semantic space, and two language-dependent decoders. In contrast, \cite{lample2018unsupervised} ensured the two languages share the same encoder and decoder, relying on an identifier to indicate specific language similar to single-model based multilingual NMT \cite{johnson2017google}. The architecture and training objective functions are illustrated in Fig.~\ref{fig12}.

The top in Fig.~\ref{fig12} shows the use of denoising auto-encoder. The encoder encodes a noisy version of the input $x$ into hidden representation $z_{src}$ which is used to reconstruct the input with the decoder. The distance (auto-encoder loss $\mathcal{L}_{auto}$) between the reconstruction $x'$ and the input $x$ should be as small as possible. To guarantee source and target languages share the same semantic space, an adversarial loss $\mathcal{L}_{adv}$ is introduced to fool the language identifier.

The bottom in Fig.~\ref{fig12} illustrates the use of back-translation. A target sentence $y$ is first back-translated into $x^*$ using an old target-to-source NMT model (the one optimized in previous iteration, and the initial model is the word-by-word translation model based on bilingual word induction). Then, the noisy version of the translation $x^*$ is encoded into $z_{src}$ which is then translated back into target sentence $y'$. The new NMT model (encoder and decoder) is optimized to minimize the translation loss $\mathcal{L}_{trans}$ which is the distance between the translation $y'$ and the original target input $y$. Similarly, an adversarial loss is employed in the encoder module. This process iterates until convergence of the algorithm. Finally, the encoder and decoder can be applied to perform dual translation tasks.

\cite{yang2018unsupervised} argued that sharing some layers of encoder and decoder while making others language-specific could improve the performance of unsupervised NMT. \cite{artetxe2019effective} further combined the NMT and SMT to improve the unsupervised translation quality. Most recently, \cite{conneau2019cross,song2019mass,ren2019explicit} resorted to pre-training techniques to enhance the unsupervised NMT model. For example, \cite{conneau2019cross} proposed a cross-lingual language model pre-training method under BERT framework \cite{devlin2019bert}. Then, two pre-trained cross-lingual language models are employed as the encoder and decoder respectively to perform translation.

\section{Multimodal neural machine translation}\label{section7}

\begin{figure*}[t]
	\centering
	\includegraphics[scale=.6]{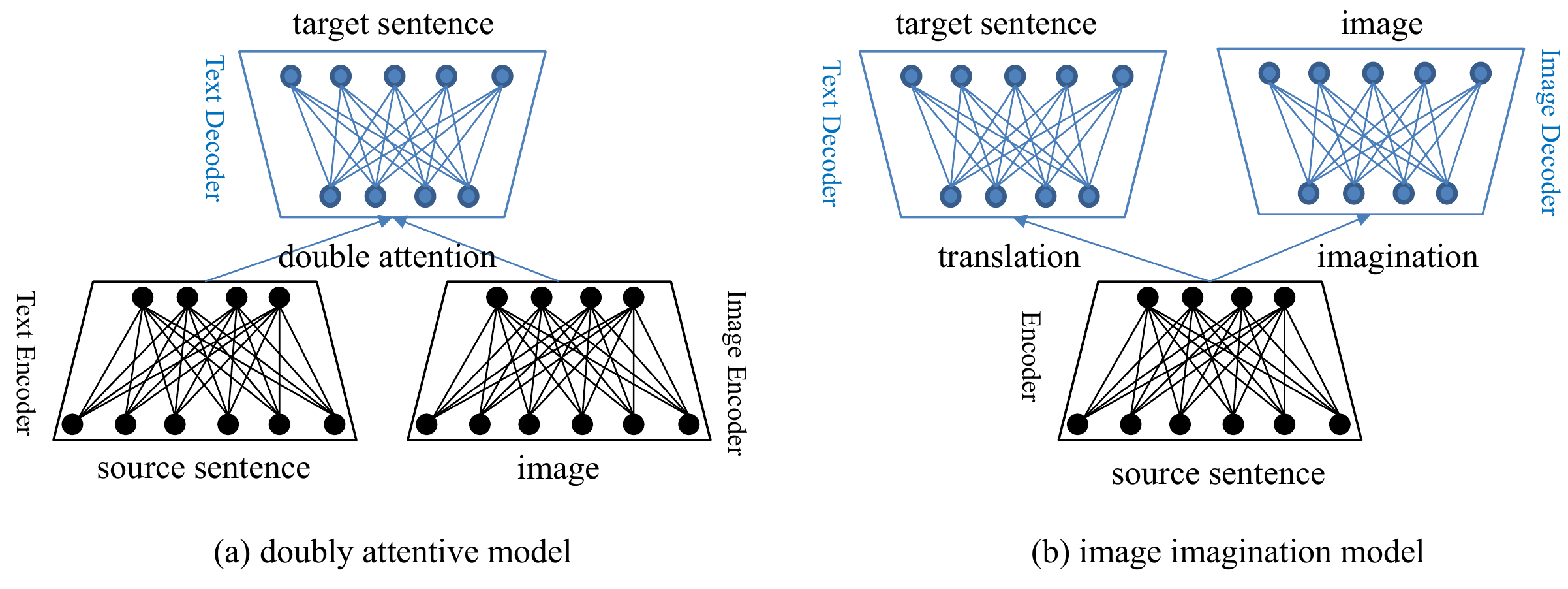}
	\caption{comparison between the doubly-attentive model and the image imagination model for image-text translation. In doubly attentive model, the image is encoded as an additional input feature. While in image imagination model, the image is decoded output from the source sentence.}
	\label{fig3}
\end{figure*}

We know that humans communicate with each other in a multimodal environment in which we see, hear, smell and so on. Naturally, it is ideal to perform machine translation with the help of texts, speeches and images. Unfortunately, video corpora containing parallel texts, speech and images for machine translation are not publicly available currently. Recently, IWSLT-2020{\footnote{http://iwslt.org/doku.php?id=evaluation}} organized the first evaluation on vedio translation in which annotated video data is only available for validation and test sets.

Translation for paired image-text, offline speech-to-text translation and simultaneous translation have become increasingly popular in recent years.

\subsection{Image-Text Translation}\label{section71}

Given an image and its text description as source language, the task of image-text translation aims at translating the description in source language into the target language, where the translation process can be supported by information from the paired image. It is a task requiring the integration of natural language processing and computer vision. WMT{\footnote{https://www.statmt.org/wmt16/multimodal-task.html}} organized the first evaluation task on image-text translation (they call it multimodal translation) in 2016 and also released the widely-used dataset {\emph{Multi30K}} consisting of about 30K images each of which has an English description and translations in German, French and Czech{\footnote{https://github.com/multi30k/dataset}}. Several effective models have been proposed since then. These methods mainly differ in the usage of the image information and we mainly discuss four of them in this section.

\cite{huang2016attention} proposed to encode the image into one distributed vector representation or a sequence of vector representations using convolutional neural networks. Then, they padded the vector representations together with the sentence as the final input for the NMT model which does not need to be modified for adaptation. The core idea is that they did not distinguish images from texts in the model design.

\cite{calixto2017doubly} presented a doubly-attentive decoder for the image-text translation task. The major difference from \cite{huang2016attention} is that they design textual encoder and visual encoder respectively, and employ two separate attention models to balance the contribution of text and image during prediction at each time-step.

\cite{elliott2017imagination} introduced a multi-task learning framework to perform image-text translation. They believe that one can imagine the image given the source language sentence. Based on this assumption, they use one encoder and two decoders in a multi-task learning framework. The encoder first encodes the source sentence into distributed semantic representations. One decoder generates the target language sentence from the source-side representations. The other decoder is required to reconstruct the given image. It is easy to see that the images are only employed in the training stage but are not required during testing. From this perspective, the multi-task learning framework can be applicable in more practical scenarios.

\cite{calixto2019latent} further proposed a latent variable model for image-text translation. Different from previous methods, they designed a generative model in which a latent variable is in charge of generating the source and target language sentences, and the image as well.

Fig.~\ref{fig3} illustrates the comparison between the doubly-attentive model and the image imagination model. Suppose the paired training data is $D=\{({\bm{x}}^{(m)},{\bm{y}}^{(m)},{IM}^{(m)})\}_{m=1}^M$ where $IM$ denotes image. The objective function of the doubly-attentive model can be formulated as follows:

\begin{equation}
\mathcal{L}(\theta) = \sum_{m=1}^{M} {\text{log}}P({\bm y}^{(m)}|{\bm x}^{(m)}, {IM}^{(m)};\theta)
\end{equation}

In contrast, the image imagination model has the following objective function which includes two parts, one for text translation and the other for image imagination.

\begin{equation}
\mathcal{L}(\theta) = \sum_{m=1}^{M}\bigg( {\text{log}}P({\bm y}^{(m)}|{\bm x}^{(m)};\theta) + {\text{log}}P({IM}^{(m)}|{\bm x}^{(m)};\theta) \bigg)
\end{equation}

All the above methods are proved to significantly improve the translation quality. But it remains a natural question that when and how does the image help the text translation. \cite{calixto2019error} conducted a detailed error analysis when translating both visual and non-visual terms. They find that almost all kinds of translation errors (not only the terms having strong visual connections) have decreased after using image as the additional context.

Alternatively, \cite{caglayan2019probing} attempted to answer when the visual information is needed in the image-text translation. They designed an input degradation method to mask crucial information in the source sentence (e.g. masking color words or entities) in order to see whether the paired image would make up the missing information during translation. They find that visual information of the image can be helpful when it is complementary rather than redundant to the text modality.

\subsection{Offline Speech-to-Text Translation}\label{section72}

Speech-to-Text translation abbreviated as speech translation (ST) is a task that automatically converts the speech in the source language (e.g. English) into the text in the target language (e.g. Chinese). Offline speech translation indicates that the complete speech (e.g. a sentence or a fragment in a time interval) is given before we begin translating. Typically, ST is accomplished with two cascaded components. Source language speech is first transcribed into the source language text using an automatic speech recognition (ASR) system. Then, the transcription is translated into target language text with a neural machine translation system. It is still the mainstream approach to ST in real applications. In this kind of paradigm, ASR and NMT are not coupled and can be optimized independently.

Nevertheless, the pipeline method has two disadvantages. On one hand, the errors propagate through the pipeline and the ASR errors are difficult to make up during translation. On the other hand, the efficiency is limited due to the two-phase process. \cite{zong1999analysis} believed in early years that end-to-end speech translation is possible with the development of memory, computational capacity and representation models. Deep learning based on distributed representations facilitates the end-to-end modeling for speech translation. \cite{berard2016listen} presented an end-to-end model without using any source language transcriptions under an encoder-decoder framework. Different from the pipeline paradigm, the end-to-end model should be optimized on the training data consisting of instances (source speech, target text).  We list some of the recently used datasets in Table.~\ref{st-datasets}, including IWSLT{\footnote{http://i13pc106.ira.uka.de/~mmueller/iwslt-corpus.zip}}, Augmented Librispeech{\footnote{https://persyval-platform.univ-grenoble-alpes.fr/DS91/detaildataset}}, Fisher and Callhome{\footnote{https://github.com/joshua-decoder/fisher-callhome-corpus}}, MuST-C{\footnote{mustc.fbk.eu}} and TED-Corpus{\footnote{https://drive.google.com/drive/folders/1sFe6Qht4vGD49vs7\_gbrNEOsLPOX9VIn?usp=sharing}}.

\begin{table*}
	\centering
	\begin{tabular}{c|l|l|l|l}
		\hline
		Corpus Name                    & Source Language & Target Language & Hours & Sentents   \\
		\hline
		\hline
		IWSLT \cite{jan2018iwslt} & En & De & 273 & 171,121 \\
		Augmented Librispeech & En & Fr & 236 & 131,395 \\
		Fisher and Callhome \cite{post2013improved}  & En & Es & 160 & 153,899 \\
		MuST-C \cite{di2019must} & En & De, Es, Fr, It, Nl, Pt, Ro, Ru & 385-504 & 4.0M-5.3M \\
		TED-Corpus \cite{liu2020synchronous} & En & De, Fr, Zh, Ja & 520 & 235K-299K \\
		\hline
	\end{tabular}
	\caption{some datasets used in the end-to-end ST. En, De, Fr, Es, It, Nl, Pt, Ro, Ru, Zh and Ja denote English, German, French, Spanish, Italian, Dutch, Portuguese, Romanian, Russian, Chinese and Japanese respectively.} \label{st-datasets}
\end{table*}

It is easy to find from the table that the training data for end-to-end ST is much less than that in ASR and MT. Accordingly, most of recent studies focus on fully utilizing the data or models of ASR and NMT to boost the performance of ST. Multi-task learning \cite{weiss2017sequence,anastasopoulos2018tied,berard2018end}, knowledge distillation \cite{jia2019leveraging,liu2019end} and pre-training \cite{bansal2019pre,wang2020bridging} are three main research directions.

In the multi-task learning framework, the ST model is jointly trained with the ASR and MT models. Since the ASR and MT models are optimized on massive training data, the ST model can be substantially improved through sharing encoder with the ASR model and decoder with the MT model. \cite{weiss2017sequence} showed that great improvements can be achieved under multi-task learning. While \cite{berard2018end} demonstrated that multi-task learning could also accelerate the convergence in addition to better translation quality.

In contrast to the multi-task learning framework, the pre-training method first pre-trains an ASR model or an MT model, then the encoder of ASR or the decoder of MT can be utilized to directly initialize the components of the ST model. \cite{bansal2019pre} attempted to pre-train ST model with the ASR data to promote the acoustic model and showed that pre-training a speech encoder on one language can boost the translation quality of ST on a different source language. To further bridge the gap between pre-training and fine-tuning, \cite{wang2020bridging} only pre-trained the ASR encoder to maximize Connectionist Temporal Classification (CTC) objective function \cite{graves2006connectionist}. Then, they share the projection matrix between the CTC classification layer for ASR and the word embeddings. The text sentence in MT is converted into the same length as the CTC output sequence of the ASR model. By doing this,  the ASR encoder and the MT encoder will be consistent in length and semantic representations. Therefore, the pre-trained encoder and attention in the MT model can be used in ST in addition to the ASR encoder and the MT decoder.

\begin{figure}[H]
	\centering
	\includegraphics[scale=.9]{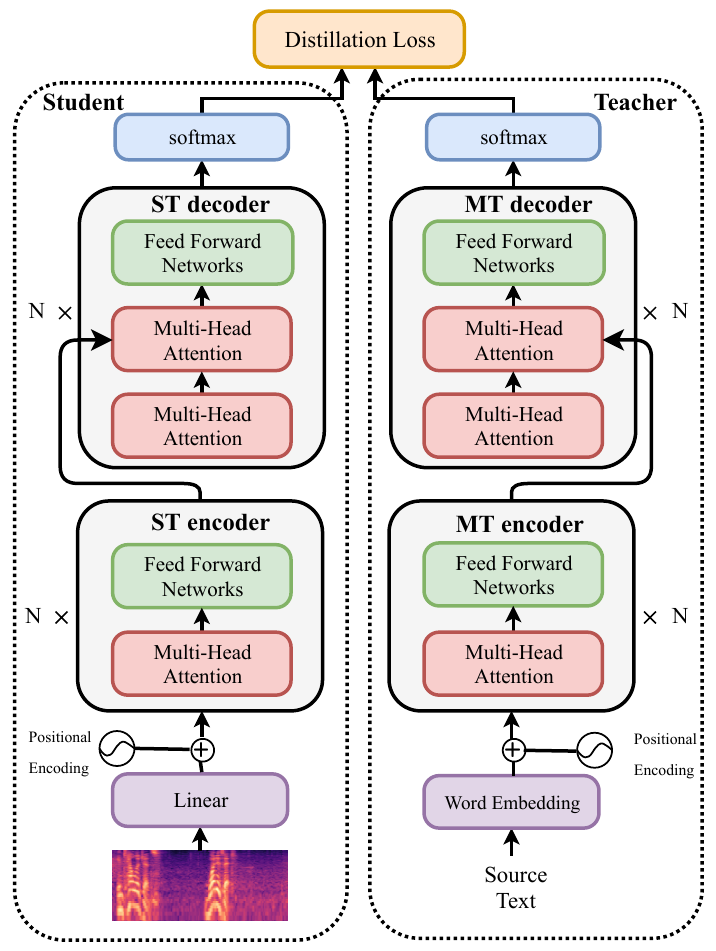}
	\caption{the illustration of the knowledge distillation model for ST. The right part is an MT model, which is a teacher. The left part is the ST model which is the student. The input of the ST model is raw speech and the input of the MT model is the transcription of the speech. The distillation loss in the top part makes the student model learn output probability distributions from the teacher model (mimic the behavior of the teacher).}
	\label{fig4}
\end{figure}

Different from the multi-task learning framework and the pre-training methods which attempt to share network parameters among ASR, ST and MT, the knowledge distillation methods consider the ST model as a student and make it learn from the teacher (e.g. the MT model). \cite{liu2019end} proposed the knowledge distillation model as shown in Fig.~\ref{fig4}. Given the training data of ST $D=\{({\bm{s}}^{(m)},{\bm{x}}^{(m)},{\bm{y}}^{(m)})\}_{m=1}^M$, where $\bm{s}$ denotes the speech segment, $\bm{x}$ is the transcription in source language and $\bm{y}$ is the translation text in target language.

The objective function for ST is similar to MT and the only difference is that the input is speech segment rather than a text sentence.

\begin{equation}
\mathcal{L}_{ST}(\theta) = - \sum_{(\bm{s}, \bm{y})\in D} {\text{log}}P({\bm y}^{(m)}|{\bm s}^{(m)};\theta)
\end{equation}

\begin{equation}
{\text{log}}P({\bm y}|{\bm s},\theta)=\sum_{i=0}^{I}\sum_{v=1}^{|V|}\mathbb{I}(y_i=v){\text{log}}P(y_i|{\bm s},{\bm y}_{<i},\theta)
\end{equation}
where $|V|$ denotes the vocabulary size of the target language and $\mathbb{I}(y_i=v)$ is the indication function which indicates whether the $i$-th output token $y_i$ happens to be the ground truth.

Given the MT teacher model pre-trained on large-scale data, it can be used to force decode the pair $({\bm{x}}, {\bm{y}})$ from the triple $({\bm{s}}, {\bm{x}}, {\bm{y}})$ and will obtain a probability distribution for each target word $y_i$: $Q(y_i|{\bm{x}}, {\bm{y}}_{<i}; \theta_{MT})$. Then, the knowledge distillation loss can be written as follows:

\begin{equation}
\mathcal{L}_{KD}(\theta) = - \sum_{(\bm{x}, \bm{y})\in D} \sum_{i=0}^{I}\sum_{v=1}^{|V|}Q(y_i|{\bm{x}}, {\bm{y}}_{<i}; \theta_{MT}){\text{log}}P(y_i|{\bm x},{\bm y}_{<i},\theta)
\end{equation}

The final ST model can be trained by optimizing both of the log-likelihood loss $\mathcal{L}_{ST}(\theta)$ and the knowledge distillation loss $\mathcal{L}_{KD}(\theta)$.

\begin{figure}[H]
	\centering
	\includegraphics[scale=.7]{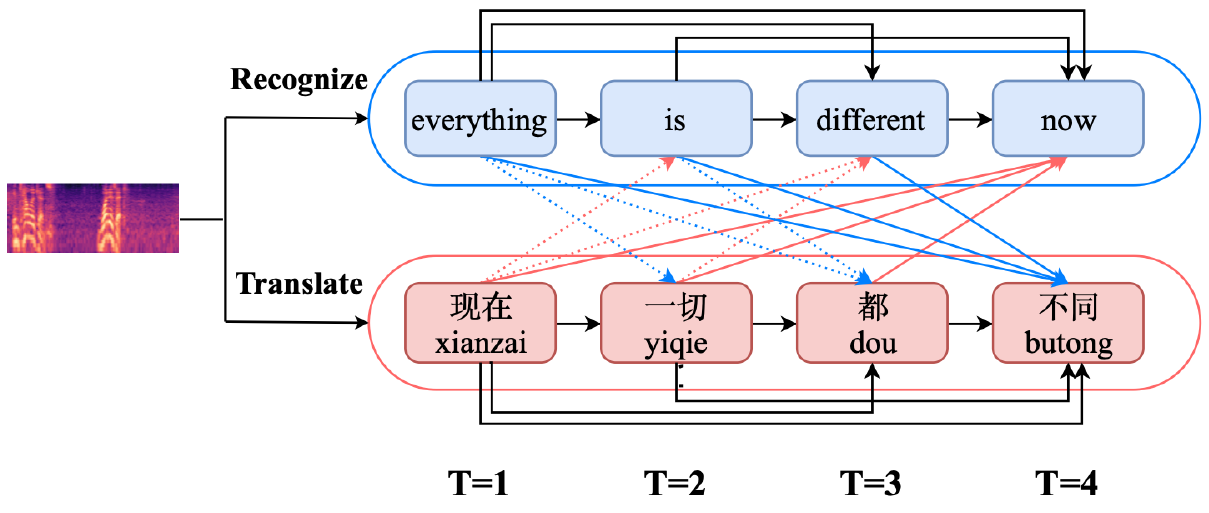}
	\caption{the demonstration of the interactive model for both ASR and ST. Taking $T=2$ as an example, the transcription "everything" of the ASR model can be helpful to predict the Chinese translation at $T=2$. Likewise, the translation at time step $T=1$ is also beneficial to generate the transcriptions of the ASR model in the future time steps.}
	\label{fig5}
\end{figure}

In order to fully explore the integration of ASR and ST, \cite{liu2020synchronous} further proposed an interactive model in which ASR and MT perform synchronous decoding. As shown in Fig.~\ref{fig5}, the dynamic outputs of each model can be used as the context to improve the prediction of the other model. Through interaction, the quality of both models can be significantly improved while keeping the efficiency as much as possible.

\subsection{Simultaneous Machine Translation}\label{section73}

\begin{figure*}[t]
	\centering
	\includegraphics[scale=.5]{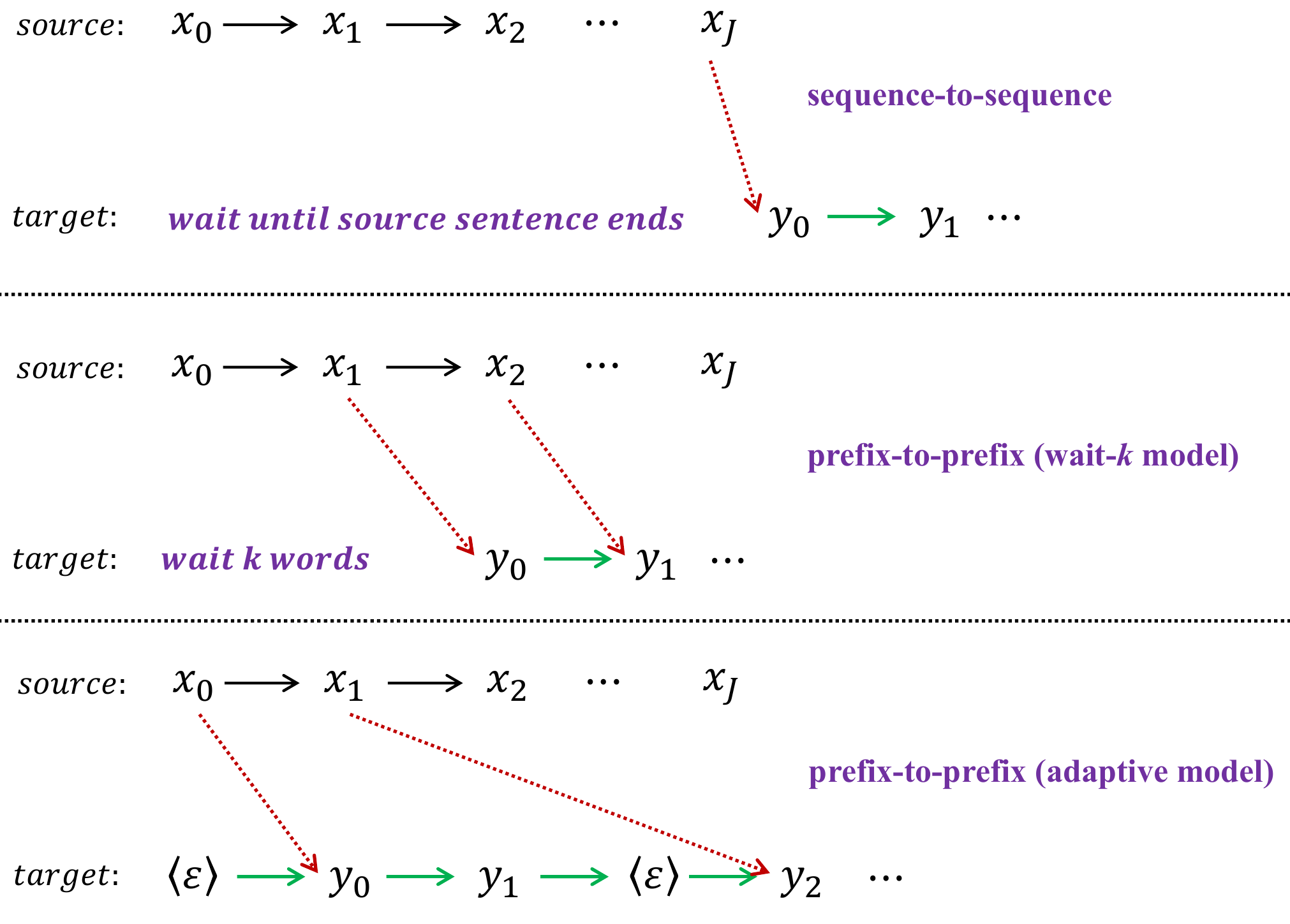}
	\caption{the illustration of three policies for simultaneous machine translation. The top part is the conventional sequence-to-sequence MT model which begins translation after seeing the whole source sentence. The middle one demonstrates the wait-$k$ policy which waits for $k$ words before translation. The bottom part shows an example of the adaptive policy that predicts an output token at each time step. If the output is a special token $\langle\varepsilon\rangle$, it indicates reading one more source word.}
	\label{fig6}
\end{figure*}

Simultaneous machine translation (SimMT) aims at translating concurrently with the source-language speaker speaking. It addresses the problem where we need to incrementally produce the translation while the source-language speech is being received. This technology is very helpful for live events and real-time video-call translation. Recently, Baidu and Facebook organized the first evaluation task on SimMT in ACL-2020{\footnote{https://autosimtrans.github.io/shared}} and IWSLT-2020{\footnote{http://iwslt.org/doku.php?id=simultaneous\_translation}} respectively.

Obviously, the methods of offline speech translation introduced in the previous section cannot be applicable in these scenarios, since the latency must be intolerable if translation begins after speakers complete their utterance. Thus, balancing between latency and quality becomes the key challenge for the SimMT system. If it translates before the necessary information arrives, the quality will decrease. However, the delay will be unnecessary if it waits for too much source-language contents.

\cite{niehues2016dynamic,niehues2018low} proposed to directly perform simultaneous speech-to-text translation, in which the model is required to generate the target-language translation from the incrementally incoming foreign speech. In contrast, more research work focus on the simultaneous text-to-text translation where they assume that the transcriptions are correct \cite{grissom2014don,satija2016simultaneous,gu2017learning,alinejad2018prediction,dalvi2018incremental,ma2019stacl,arivazhagan2019monotonic,zheng2019simpler,zheng2019simultaneous,arthur2020learning}. This article mainly introduces the latter methods. All of the methods address the same strategy (also known as policy) that when to read an input word from the source language and when to write an output word in target language, namely when to wait and when to translate.

In general, the policies can be categorized into two bins. One is fixed-latency policies \cite{dalvi2018incremental,ma2019stacl}, such as wait-$k$ policy. The other is adaptive policies \cite{satija2016simultaneous,gu2017learning,arivazhagan2019monotonic,zheng2019simpler,zheng2019simultaneous,arthur2020learning}. 

The wait-$k$ policy proposed by \cite{ma2019stacl} is proved simple but effective. Just as shown in the middle part of Fig.~\ref{fig6}, the wait-$k$ policy starts to translate after reading the first $k$ source words. Then, it alternates between generating a target-language word and reading a new source-language word, until it meets the end of the source sentence. Accordingly, the probability of a target word $y_i$ is conditioned on the history predictions $\bm{y}_{<i}$ and the prefix of the source sentence $\bm{x}_{<i+k}$: $P(y_i|\bm{y}_{<i}, \bm{x}_{<i+k}; \theta)$. The probability of the whole target sentence becomes:

\begin{equation}
P(\bm{y}|\bm{x}; \theta)=\prod_{i=0}^{I}P(y_i|\bm{y}_{<i}, \bm{x}_{<i+k}; \theta)
\end{equation}

In contrast to previous sequence-to-sequence NMT training paradigm, \cite{ma2019stacl} designed a prefix-to-prefix training style to best explore the wait-$k$ policy. If Transformer is employed as the basic architecture, prefix-to-prefix training algorithm only needs a slight modification. The key difference from Transformer is that prefix-to-prefix model conditions on the first $i+k$ rather than all source words at each time-step $i$. It can be easily accomplished by applying the masked self-attention during encoding the source sentence. In that case, each source word is constrained to only attend to its predecessors and the hidden semantic representation of the $i+k$-th position will summarize the semantics of the prefix $\bm{x}_{<i+k}$. 

However, the wait-$k$ policy is a fixed-latency model and it is difficult to decide $k$ for different sentences, domains and languages. Thus, adaptive policy is more appealing. Early attempts for adaptive policy are based on reinforcement learning (RL) method. For example, \cite{gu2017learning} presented a two-stage model that employs the pre-trained sentence-based NMT as the base model. On top of the base model, read or translate actions determine whether to receive a new source word or output a target word. These actions are trained using the RL method by fixing the base NMT model.

Differently, \cite{zheng2019simpler} proposed an end-to-end simMT model for adaptive policies. They first add a special {\emph{delay}} token $\langle\varepsilon\rangle$ into the target-language vocabulary. As shown in the bottom part of Fig.~\ref{fig6}, if the model predicts $\langle\varepsilon\rangle$, it needs to receive a new source word. To train the adaptive policy model, they design dynamic action oracles with aggressive and conservative bounds as the expert policy for imitation learning. Suppose the prefix pair is $(\bm{s},\bm{t})$. Then, the dynamic action oracle can be defined as follows:

\begin{equation*} 
\Pi_{\bm{x}, \bm{y}, \alpha, \beta}^\star(\bm{s},\bm{t}) =
\begin{cases}
\{\langle \varepsilon \rangle\} & \text{if $\bm{s}\neq \bm{x}$ and $|\bm{s}|-|\bm{t}|\leq \alpha$}\\ \{y_{|\bm{t}|+1}\} & \text{if $\bm{t}\neq \bm{y}$ and $|\bm{s}|-|\bm{t}|\geq \beta$}  \\ \{\langle\varepsilon\rangle,y_{|\bm{t}|+1}\} & \text{otherwise}  
\end{cases}
\end{equation*}
where $\alpha$ and $\beta$ are hyper-parameters, denoting aggressive and conservative bounds respectively. $|\bm{s}|-|\bm{t}|$ calculates the distance between two prefixes. That is to say if the current target prefix $\bm{t}$ is no more than $\alpha$ words behind the source prefix $\bm{s}$, we can read a new source word. If $\bm{t}$ is shorter than $\bm{s}$ with more than $\beta$ words, we generate the next target prediction.

\section{Discussion and Future Research Tasks}\label{section8}
\subsection{NMT vs. Human}\label{section81}
We can see from Sec.\ref{section4}-\ref{section7} that great progresses have been achieved in neural machine translation. Naturally, we may wonder whether current strong NMT systems could perform on par with or better than human translators. Exciting news were reported in 2018 by \cite{hassan2018achieving} that they achieved human-machine parity on Chinese-to-English news translation and they found no significant difference of human ratings between their MT outputs and professional human translations. Moreover, the best English-to-Czech system submitted to WMT 2018 by \cite{popel2018cuni} was also found to perform significantly better than the human-generated reference translations \cite{bojar2018findings}. It is encouraging that NMT can achieve very good translations in some specific scenarios and it seems that NMT has achieved the human-level translation quality.

However, we cannot be too optimistic since the MT technology is far from satisfactory. On one hand, the comparisons were conducted only on news domain in specific language pairs where massive parallel corpora are available. In practice, NMT performs quite poorly in many domains and language pairs, especially for the low-resource scenarios such as Chinese-Hindi translation. On the other hand, the evaluation methods on the assessment of human-machine parity conducted by \cite{hassan2018achieving} should be much improved as pointed out by \cite{laubli2020set}. According to the comprehensive investigations conducted by \cite{laubli2020set}, human translations are much preferred over MT outputs if using better rating techniques, such as choosing professional translators as raters, evaluating documents rather than individual sentences and utilizing original source texts instead of source texts translated from target language. Current NMT systems still suffer from serious translation errors of mistranslated words or named entities, omissions and wrong word order. Obviously, there is much room for NMT to improve and we suggest some potential research directions in the next section.

\subsection{Future Research Tasks}\label{section82}
In this section, we discuss some potential research directions for neural machine translation.

\textbf{1. Effective document-level translation and evaluation} 

It is well known that document translation is important and the current research results are not so good. It remains unclear what is the best scope of the document context needed to translate a sentence. It is still a question whether it is reasonable to accomplish document translation by translating the sentences from first to last. Maybe translation based on sentence group is a worthy research topic which models many-to-many translation. In addition, document-level evaluation is as important as the document-level MT methods, and it serves as a booster of MT technology. \cite{hassan2018achieving} argued that MT can achieve human parity in Chinese-to-English translation on specific news tests if evaluating sentence by sentence. However, as we discussed in the previous section that \cite{laubli2018machine,laubli2020set} demonstrated a stronger preference for human over MT when evaluating on document-level rather than sentence-level. Therefore, how to automatically evaluate the quality of document translation besides BLEU \cite{papineni2002bleu} is an open question although some researchers introduce several test sets to investigate some specific discourse phenomena \cite{muller2018large}. 

\textbf{2. Efficient NMT inference}

People prefer the model with both high accuracy and efficiency. Despite of remarkable speedup, the quality degradation caused by non-atuoregressive NMT is intolerable in most cases. Improving the fertility model, word ordering of decoder input and dependency of the output will be worthy of a good study to make NAT close to AT in translation quality. Synchronous bidirectional decoding deserves deeper investigation due to good modeling of history and future contexts. Moreover, several researchers start to design decoding algorithm with free order \cite{gu2019insertion,stern2019insertion,emelianenko2019sequence} and it may be a good way to study the nature of human language generation.

\textbf{3. Making full use of multilinguality and monolingual data}

Low-resource translation is always a hot research topic since most of natural languages are lack of abundant annotated bilingual data. The potential of multilingual NMT is not fully explored and some questions remain open. For example, how to deal with data unbalance problem which is very common in multilingual NMT? How to build a good incremental multilingual NMT model for incoming new languages? Semi-supervised NMT is more practical in real applications but the effective back-translation algorithm is very time consuming. It deserves to design a much efficient semi-supervised NMT model for easy deployment. Deeply integrating pre-trained method with NMT may lead to promising improvement in the semi-supervised or unsupervised learning framework and \cite{ji2020cross,zhu2020incorporating} have already shown some good improvements. The achievements of unsupervised MT in similar language pairs (e.g. English-German and English-French) make us very optimistic. However, \cite{leng2019unsupervised} showed that unsupervised MT performs poorly on distant language pairs, obtaining no more than 10 BLEU scores in most cases. Obviously, it is challenging to design better unsupervised MT models on distant language pairs.

\textbf{4. Better exploiting multimodality in NMT}

In multimodal neural machine translation, it remains an open problem when and how to make full use of different modalities. The image-text translation only translates the image captions and is hard to be widely used in practice. It is a good research topic to find an appropriate scenario where images are indispensible during translation. In speech translation, despite of big improvement, the end-to-end framework currently cannot perform on par with the cascaded method in many cases, especially when the training data is limited \cite{liu2020synchronous}. In addition to enlarging the training data, closing the gap between different semantic spaces of ST, ASR and MT is worthy of further exploration. As for simultaneous translation, it is still on the early stage of research and many practical issues such as repetition and correction in speech are unexplored. Moreover, combining summarization and translation may be a good research direction that provides the audiences the gist of the speaker's speech with low latency.

\textbf{5. NMT with background modeling}

In many cases, machine translation is not about texts, speeches and images, but is highly related to culture, environment, history and etc. Accordingly, this kind of background information should be well captured by a novel model which guides NMT to generate translations in line with the background.

\textbf{6. Incorporating prior knowledge into NMT}

Note that some research topics are not mentioned in this article due to space limit. For example, it is a difficult and practical issue how to integrate prior knowledge (e.g. alignment, bilingual lexicon, phrase table and knowledge graphs) into the NMT framework. Since it is still unclear how to bridge discrete symbol based knowledge and distributed representation based NMT, it remains an interesting and important direction to explore although some progress has been achieved \cite{zhang2017prior,tu2016modeling,mi2016coverage,feng2017memory,zhao2018phrase,zhao2018addressing,wang2017translating,wang2018incorporating,lu2018exploiting}.

\textbf{7. Better domain adaption models}

Domain adaptation has been always a hot research topic and attracts attentions from many researchers \cite{luong2015stanford,zoph2016transfer,wang2017instance,chu2017empirical,chu2018survey,li2018one,zhang2019curriculum,zeng2019iterative}. Different from methods used in SMT, domain adaptation in NMT is usually highly related with parameter fine-tuning. It remains a challenge how to address the problem of unknown test domain and out-of-domain term translations.

\textbf{8. Bridging the gap between training and inference}

The inconsistency between training and inference (or evaluation) is a critical problem in most sequence generation tasks in addition to neural machine translation. This problem is well addressed in the community of machine translation \cite{shen2016minimum,zhang2019bridging} but it is still worthy of exploring especially on the efficiency of the methods.

\textbf{9. Designing explainable and robust NMT}

So far, the NMT model is still a black box and it is very risky to use it in many scenarios in which we have to know how and why the translation result is obtained. \cite{ding2017visualizing} attempted to visualize the contribution of each input to the output translation. Nevertheless, it will be great to deeply investigate the explanation of the NMT models or design an explainable MT architecture. Furthermore, current NMT systems are easy to attack through perturbing the input. \cite{cheng2018towards,cheng2019robust} presented novel robust NMT models to handle noisy inputs. However, the real input noise is too difficult to anticipate and it still remains a big challenge to design robust NMT models  which are immune to real noise.

\textbf{10. New NMT architectures} 

Finally, designing better NMT architectures beyond Transformer must be very exciting to explore despite of the difficulty.

%%%%%%%%%%%%%%%%%%%%%%%%%%%%%%%%%%%%%%%%%%%%%%%%%%%%%%%
%%% Acknowledgements. ??л
%%%%%%%%%%%%%%%%%%%%%%%%%%%%%%%%%%%%%%%%%%%%%%%%%%%%%%%
%\Acknowledgements{This work was supported by the National Natural Science Foundation of China (Grant Nos. 61234003, 61434004, 61504141) and CAS Interdisciplinary Project (Grant No. KJZD-EW-L11-04).}%

%%%%%%%%%%%%%%%%%%%%%%%%%%%%%%%%%%%%%%%%%%%%%%%%%%%%%%%
%%% Conflict of interest. ????????????
%%%%%%%%%%%%%%%%%%%%%%%%%%%%%%%%%%%%%%%%%%%%%%%%%%%%%%%
%\InterestConflict{The authors declare that they have no conflict of interest.}

%%%%%%%%%%%%%%%%%%%%%%%%%%%%%%%%%%%%%%%%%%%%%%%%%%%%%%%
%%% Supplements. ????????, ????
%%%%%%%%%%%%%%%%%%%%%%%%%%%%%%%%%%%%%%%%%%%%%%%%%%%%%%%
%\Supplements{}

%%%%%%%%%%%%%%%%%%%%%%%%%%%%%%%%%%%%%%%%%%%%%%%%%%%%%%%
%%% Reference section. ?ο?????
%%% citation in the content using "some words~\cite{1,2}".
%%% ~ is needed to make the reference number is on the same line with the word before it.
%%%%%%%%%%%%%%%%%%%%%%%%%%%%%%%%%%%%%%%%%%%%%%%%%%%%%%%
\section*{Reference}
\bibliographystyle{plain}
\bibliography{mybibfile}

%%%%%%%%%%%%%%%%%%%%%%%%%%%%%%%%%%%%%%%%%%%%%%%%%%%%%%%
%%% Appendix sections. ??????, ????
%%%%%%%%%%%%%%%%%%%%%%%%%%%%%%%%%%%%%%%%%%%%%%%%%%%%%%%
%\begin{appendix}

%\renewcommand{\thesection}{Appendix}%不加ABC

%\section{}

%\end{appendix}

\end{multicols}

\end{document}